\documentclass[twoside]{article}

\usepackage[accepted]{aistats2020}
\usepackage[utf8]{inputenc} % allow utf-8 input
\usepackage[T1]{fontenc}    % use 8-bit T1 fonts
\usepackage{hyperref}       % hyperlinks
\usepackage{url}            % simple URL typesetting
\usepackage{booktabs}       % professional-quality tables
\usepackage{amsfonts}       % blackboard math symbols
\usepackage{nicefrac}       % compact symbols for 1/2, etc.
\usepackage{microtype}      % microtypography
\usepackage{nicefrac}
\usepackage{dsfont}
\usepackage{wrapfig}
\usepackage{amsmath,amsfonts,amssymb}

\usepackage{times}  %Required
\usepackage{helvet}  %Required
\usepackage{courier}  %Required
\usepackage{url}  %Required
\usepackage{graphicx}  %Required
\usepackage{xfrac}
\usepackage{xcolor}
\usepackage{mathtools}
\usepackage{lipsum}
\usepackage{bigints}
\usepackage{bbm}
\usepackage{multirow}
\usepackage{array}
\usepackage{textcomp}
\usepackage{xcolor}
\usepackage{subcaption}
\usepackage{natbib}

\DeclareMathOperator*{\argmax}{arg\,max}
\newtheorem{theorem}{Theorem}

\newtheorem{remark}{Remark}

\DeclareMathOperator{\E}{\mathbb{E}}

\hypersetup{draft}

\begin{document}

\twocolumn[

\aistatstitle{Optimization Methods for Interpretable Differentiable Decision Trees in Reinforcement Learning}

% \aistatsauthor{ Author 1 \And Author 2 \And  Author 3 }

\aistatsauthor{ Andrew Silva \And Taylor Killian \And Ivan Rodriguez Jimenez}
\aistatsaddress{ Georgia Institute of Technology \And University of Toronto \And Georgia Institute of Technology}
\aistatsauthor{Sung-Hyun Son \And Matthew Gombolay}
\aistatsaddress{ MIT Lincoln Laboratory \And Georgia Institute of Technology}
]

\begin{abstract}
 Decision trees are ubiquitous in machine learning for their ease of use and interpretability. Yet, these models are not typically employed in reinforcement learning as they cannot be updated online via stochastic gradient descent. We overcome this limitation by allowing for a gradient update over the entire tree that improves sample complexity affords interpretable policy extraction. First, we include theoretical motivation on the need for policy-gradient learning by examining the properties of gradient descent over differentiable decision trees. Second, we demonstrate that {{our approach equals or outperforms a neural network on all domains and can learn discrete decision trees online with average rewards up to 7x higher than a batch-trained decision tree.}} Third, we conduct a user study to quantify the interpretability of a decision tree, rule list, and a neural network {{with statistically significant results ($p < 0.001$).}}
\end{abstract}

\runningauthor{ Andrew Silva, Taylor Killian, Ivan Rodriguez Jimenez, Sung-Hyun Son, Matthew Gombolay}

\section{Introduction and Related Work}
\label{sec:intro}

Reinforcement learning (RL) with neural network function approximators, known as ``Deep RL,'' has achieved tremendous results in recent years~\citep{Andrychowicz2018LearningDI,andrychowicz2016learning,arulkumaran2017brief,espeholt2018impala,mnih2013playing,sun2018tstarbots,rajeswaran2017learning}. Deep RL uses multi-layered neural networks to represent policies trained to maximize an agent's expected future reward. Unfortunately, these neural-network-based approaches are largely uninterpretable due to the millions of parameters involved and nonlinear activations throughout. 

In safety-critical domains, e.g., healthcare and aviation, insight into a machine's decision-making process is of utmost importance. Human operators must be able to follow step-by-step procedures~\citep{,gawande2010checklist,haynes2009surgical} or hold machines accountable \citep{natarajan2020trust}. Of the machine learning (ML) methods able to generate such procedures, decision trees are among the most highly developed~\citep{weiss1995rule}, persisting in use today~\citep{gombolay2018human,zhang2019interpreting}. While interpretable ML methods offer much promise~\citep{letham2015interpretable}, they are unable to match the performance of Deep RL~\citep{finney2002thing,silver2016mastering}. In this paper, we advance the state of the art in decision tree methods for RL and leverage their ability to yield interpretable policies.
% clay2015back

Decision trees are viewed as the de facto technique for interpretable and transparent ML~\citep{rudin2014algorithms,lipton2018mythos}, as they learn compact representations of relationships within data~\citep{breiman1984classification}. Rule  \citep{angelino2017learning,chen2017optimization} and decision lists \citep{lakkaraju2017learning,letham2015interpretable} are related architectures also used to communicate a decision-making process. Decision trees have been also applied to RL problems where they served as function approximators, representing which action to take in which state~\citep{ernst2005tree,finney2002thing,pyeatt2001decision,shah2010fuzzy}. 

The challenge for decision trees as function approximators lies in the online nature of the RL problem. The model must adapt to the non-stationary distribution of the data as the model interacts with its environment. The two primary techniques for learning through function approximation, Q-learning~\citep{watkins1989learning} and policy gradient~\citep{sutton2000policy}, rely on online training and stochastic gradient descent~\citep{bottou2010large,fletcher1963rapidly}. Standard decision trees are not amenable to gradient descent as they are a collection of non-differentiable, nested, if-then rules. As such, researchers have used non-gradient-descent-based methods for training decision trees for RL~\citep{ernst2005tree,finney2002thing,pyeatt2001decision}, e.g., greedy state aggregation, rather than seeking to update the entire model with respect to a global loss function~\citep{pyeatt2001decision}. Researchers have also attempted to use decision trees for RL by training in batch mode, completely re-learning the tree from scratch to account for the non-stationarity introduced by an improving policy~\citep{ernst2005tree}. This approach is inefficient when scaling to realistic situations and is not guaranteed to converge.
Despite these attempts, success comparable to that of modern deep learning approaches has been elusive \citep{finney2002thing}.

% Seeking to develop a decision tree formulation amenable to gradient descent, Su\'{a}rez and Lutsko formulate a continuous and fully differentiable decision tree (DDT), in which they adopt a sigmoidal representation of each decision node's splitting criterion~\citep{suarez1999globally}. Su\'{a}rez and Lutsko apply their approach to offline, supervised learning but not to RL. Researchers continue to explore continuous decision tree formulations~\citep{olaru2003complete,kontschieder2015deep}, with limited success in application to RL (e.g., \citep{shah2010fuzzy}).

In this paper, we present an novel function approximation technique for RL via differentiable decision trees (DDTs). We provide three contributions. First, we examine the properties of gradient descent over DDTs, motivating policy-gradient-based learning. To our knowledge, this is the first investigation of the optimization surfaces of Q-learning and policy gradients for DDTs. Second, we compare our method with baseline approaches on standard RL challenges, showing {that our approach parities or outperforms a neural network; further, the interpretable decision trees we discretize after training achieve an average reward up to 7x higher than a batch-learned decision tree.} Finally, we conduct a user study to compare the interpretability and usability of each method as a decision-making aid for humans, showing that discrete trees and decision lists are {perceived as more helpful ($p < 0.001$) and are objectively more efficient ($p < 0.001$) than a neural network.}

\begin{remark}[Analysis Significance]
Our approach builds upon decades of work in machine and RL; yet ours is the first to consider DDTs for online learning. 
% As such, it is important to provide an analytical basis for selecting the most helpful training methods for DDTs.
% for this novel approach (i.e., Q-learning versus PG.) 
While researchers have shown failings of Q-learning with function approximation, including for sigmoids~\citep{baird1995residual,bertsekas1996neuro,gordon1995stable,tsitsiklis1996feature}, we are unaware of analysis of Q-learning and policy gradient for our unique architecture. Our analysis provides insight regarding the best practices for training interpretable RL policies with DDTs.
\end{remark}

\section{Preliminaries}
\vspace{-0.1cm}
In this section, we review decision trees, DDTs, and RL.

\subsection{Decision Trees}
\vspace{-0.1cm}
A decision tree is a directed, acyclic graph, with nodes and edges, that takes as input an example, $x$, performs a forward recursion, and returns a label $\hat{y}$ ( Eq.~\ref{eq:yhat}-\ref{eq:basicSplit}).\par\nobreak{\parskip0pt \footnotesize \begin{align}
\hat{y}(x) &\coloneqq  T_{\eta_o}(x) \label{eq:yhat}\\
T_{\eta}(x) &\coloneqq  
\begin{cases} 
  y_{\eta}, &  \text{if leaf} \\ % \nexists\eta_\swarrow \\
  \mu_{\eta}(x)T_{\eta_\swarrow}(x) + (1-\mu_{\eta}(x))T_{\eta_\searrow}(x) & \text{o/w}
\end{cases} \label{eq:T} \\
\mu_{\eta}(x) &\coloneqq  
\begin{cases} 
  1, &  \text{if } x_{j_{\eta}} > \phi_{\eta} \\
  0, & \text{o/w}
\end{cases} \label{eq:basicSplit}
\end{align}}There are two node types: \emph{decision} and \emph{leaf} nodes, which have an outdegree of two and zero, respectively. Nodes have an indegree of one except for the root, $\eta_o$, whose indegree is zero. Decision nodes $\eta$ are represented as Boolean expressions, $\mu_{\eta}$ (Eq.~\ref{eq:basicSplit}), where $x_{j_\eta}$ and $\phi_\eta$ are the selected feature and splitting threshold for decision node $\eta$. For each decision node, the left outgoing edge is labeled ``true,'' and the right outgoing edge is labeled ``false.'' E.g., if $\mu_{\eta}$ is evaluated true, the \emph{left child} node, $\eta_\swarrow$, is considered next. The process repeats until a leaf is reached upon which the tree returns the corresponding label. The goal is to determine the best $j_{\eta}^*$, $\phi_{\eta}^*$, and $y_{\eta}$ for each node and the best structure (i.e., whether, for each $\eta$, there exists a child). There are many heuristic techniques for learning decision trees with a batch data~\citep{breiman1984classification}. However, one cannot apply gradient updates as the tree is fixed at generation. While some have sought to grow trees for RL~\citep{pyeatt2001decision}, these approaches do not update the entire tree.

\paragraph{DDTs --} Su\'{a}rez and Lutsko provide one of the first DDT models. Their method replaces the Boolean decision in Eq.~\ref{eq:basicSplit} with the sigmoid activation function shown in Eq.~\ref{eq:sigmoid-function}. This function considers a linear combination of features $x$ weighted by $\beta_{\eta}$ compared to a bias value $\phi_\eta$, and augmented by a steepness parameter $\alpha_{\eta}$. The tree is trained via gradient descent for, $\phi_{\eta}$, $\beta_{\eta}$, and $\alpha_{\eta}$ across nodes $\eta$~\citep{suarez1999globally}. This method has been applied to offline, supervised learning but not RL. 

\par\nobreak{\parskip0pt \small
\begin{align}
\mu_{\eta}(x) &\coloneqq \frac{1}{1+e^{-(\alpha_{\eta}(\beta_{\eta}^{\top} x  -   \phi_{\eta}))}} \label{eq:sigmoid-function}
\end{align}}
% We believe this model serves as a building block towards developing interpretable, machine learning models amenable to gradient descent. 
\vspace{-0.4cm}
\begin{figure*}
\centering
\begin{subfigure}{.15\textwidth}
  \centering
  \includegraphics[width=.85\linewidth,height = 3.25cm]{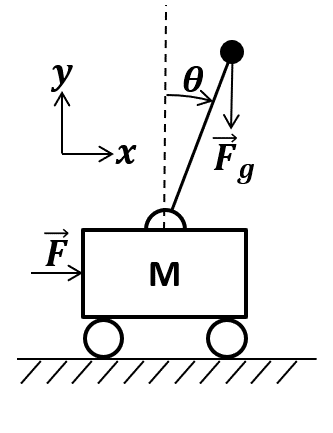}
  \caption{}
  \label{fig:CartPole}
\end{subfigure}%
\begin{subfigure}{.65\textwidth}
  \centering
  \includegraphics[width=.85\linewidth,height = 3.25cm]{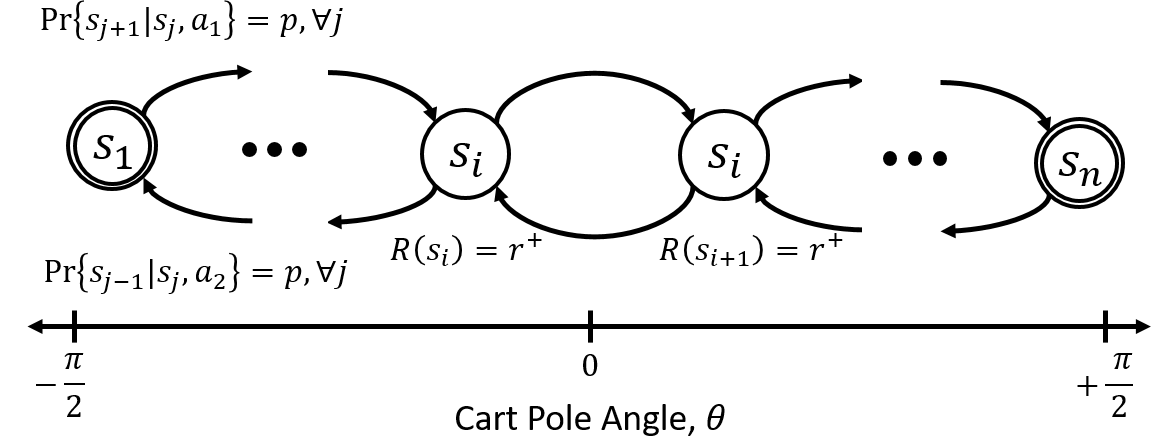}
  \caption{}
  \label{fig:MDP}
\end{subfigure}
\begin{subfigure}{.15\textwidth}
  \centering
  \includegraphics[width=.85\linewidth,height = 3.25cm]{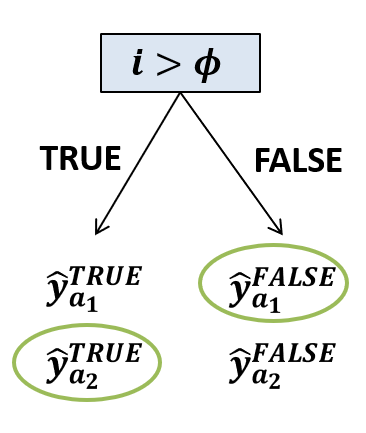}
  \caption{}
  \label{fig:Tree}
\end{subfigure}
\caption{Figure \ref{fig:CartPole} depicts the cart pole analogy for our analysis. Figure \ref{fig:MDP} depicts the MDP model for our analysis. Figure \ref{fig:Tree} depicts a tree representation of the optimal policy for our analysis, with optimal actions circled.}
\label{fig:test}
\end{figure*}
\subsection{Reinforcement Learning} RL is a subset of machine learning in which an agent is tasked with learning the optimal action sequence that maximizes future expected reward~\citep{sutton1998reinforcement}. The problem is abstracted as a Markov Decision Process (MDP), which is a five-tuple $\langle S,A,P,\gamma,R \rangle$ defined as follows: $S$ is the set of states; $A$ is the set of actions; $P: S \times A \times S \rightarrow [0,1]$ is the transition matrix describing the probability that taking action $a \in A$ in state $s \in S$ results in state $s' \in S$; $\gamma\in[0,1]$ is the discount factor defining the trade-off between immediate and future reward; and $R: S \times A \rightarrow \mathbb{R}$ is the function dictating the reward an agent receives by taking action $a \in A$ in state $s \in S$. The goal is to learn a policy, $\pi: S \rightarrow A$, that prescribes which action to take in each state to maximize the agent's long-term expected reward. There are two ubiquitous approaches to learn a policy: Q-learning and policy gradient. 

Q-learning seeks to to learn a mapping, $Q^{\pi}:S\times A \rightarrow \mathbb{R}$, that returns the expected future reward when taking action $a$ in state $s$. This mapping (i.e., the Q-function) is typically approximated by a parameterization $\theta$ (e.g., a neural network), $Q_{\theta}$. One then minimizes the Bellman residual via Eq.~\ref{eq:QLearningPolicy}, where $^{Q}\Delta \theta$ is the estimated change in $\theta$, and $s_{t+1}$ is the state the agent arrives in after applying action $a_t$ in state $s_t$ at time step $t$ with learning rate $\alpha$.\vspace{-0.2cm}\par\nobreak{\parskip0pt \small \begin{align}
^Q\Delta \theta \coloneqq \alpha &\left( R(s_t,a_t) + \gamma\max_{a' \in A} Q_{\theta}(s_{t+1},a') \right. \nonumber \\
 &- Q_{\theta}(s_t,a_t) \Big) \nabla_{\theta}Q^{\pi}_{\theta}(s_t,a_t) \label{eq:QLearningPolicy}
\end{align}}A complementary approach is the set of policy gradient methods in which one seeks to directly learn a policy, $\pi_{\theta}(s)$, parameterized by $\theta$, that maps states to actions. The update rule maximizes the expected reward of a policy, as shown in Eq.~\ref{eq:policyGradient}, where $^{PG}\Delta \theta$ indicates the change in $\theta$ for a timestep under policy gradient and $A_{t}=\sum_{t'=t}^T \gamma^{(T-t')}r_{t'}$. $T$ is the length of the trajectory. \vspace{-0.2cm}\par\nobreak{\parskip0pt \small
\begin{align}
^{PG}\Delta \theta \coloneqq \alpha \sum_{t} A_{t} \nabla_{\theta} \log \left({\pi}_{\theta}\left(s_t,a_t\right)\right)\label{eq:policyGradient}
\end{align}}We provide an investigation into the behavior of $^{Q}\Delta \theta$ and $^{PG}\Delta \theta$ as for DDTs in Section \ref{sec:analytical}.

\section{DDTs as Interpretable Function Approximators}

In this section, we derive the Q-learning and policy gradient updates for DDTs as function approximators in RL. Due to space considerations, we show the simple case of a DDT with a single decision node and two leaves with one feature $s$ with feature coefficient $\beta$, as shown in Eq.~\ref{eq:simpleTree} with the gradient shown in Equations \ref{eq:simpleTreeGrad1}-\ref{eq:simpleTreeGradn}.
% example decision tree is depicted in \ref{fig:Tree} and defined as $\pi_T$ in Eq.~\ref{eq:simpleTree}. 
% This decision tree partitions the state space into two subsets of states: those with an index, $i$, less than or equal to $\phi$ and those with an index, $i$, greater than $\phi$. 
\par\nobreak{\parskip0pt \small
\begin{align}
&f_T(s,a) = \mu(s) \hat{y}^{\text{TRUE}}_a + \left(1-\mu(s)\right) \hat{y}^{\text{FALSE}}_a
\label{eq:simpleTree} \\
&\nabla f_T(s,a) = \left[\frac{\partial f_T}{\partial \hat{y}_a^{TRUE}}, \frac{\partial f_T}{\partial \hat{y}_a^{FALSE}},\frac{\partial f_T}{\partial \alpha},\frac{\partial f_T}{\partial \beta},\frac{\partial f_T}{\partial \phi}\right]^\intercal \label{eq:simpleTreeGrad1} \\
&\frac{\partial f_T}{\partial \hat{y}_a^{TRUE}} = 1-\frac{\partial f_T}{\partial \hat{y}_a^{FALSE}} = \mu(s) \\
&\frac{\partial f_T}{\partial \alpha} = (\hat{q}_a^{TRUE}-\hat{q}_a^{FALSE})\mu(s)(1-\mu(s))(\beta s - \phi) \\
&\frac{\partial f_T}{\partial \beta} = (\hat{q}_a^{TRUE}-\hat{q}_a^{FALSE})\mu(s)(1-\mu(s))(a)(s) 
\end{align}
\begin{align}
&\frac{\partial f_T}{\partial \phi} = (\hat{q}_a^{TRUE}-\hat{q}_a^{FALSE})\mu(s)(1-\mu(s))(a)(-1) \label{eq:simpleTreeGradn}\end{align}}When utilizing a DDT as a function approximator for Q-learning, each leaf node returns an estimate of the expected future reward (i.e., the Q-value) for applying each action when in the portion of the state space dictated by the criterion of it's parent node (Eq.~\ref{eq:QTree}). 
\par\nobreak{\parskip0pt \small
\begin{align}
f_T(s,a) \rightarrow Q(s,a) = \mu(s) \hat{q}^{\text{TRUE}}_a + \left(1-\mu(s)\right) \hat{q}^{\text{FALSE}}_a \label{eq:QTree}
\end{align}}Likewise, when leveraging policy gradient methods for RL with DDT function approximation, the leaves represent an estimate of the optimal probability distribution over actions the RL agent should take to maximize its future expected reward. Therefore, the values at these leaves represent the probability of selecting the corresponding action (Eq.~\ref{eq:PGTree}). We impose the constraint that the probabilities of all actions sum to one ({\small $\hat{y}^{TRUE}_{a_1} + \hat{y}^{TRUE}_{a_2} = 1$}). 
\par\nobreak{\parskip0pt \small
\begin{align}
f_T(s,a) \rightarrow \pi(s,a) = \mu(s) \hat{\pi}^{\text{TRUE}}_a + \left(1-\mu(s)\right) \hat{\pi}^{\text{FALSE}}_a \label{eq:PGTree}
\end{align}}

% In other words, $Q(s,a) = \hat{y}_a^{TRUE}$ when $\mu(s) = 1$, $Q(s,a) = \hat{y}_a^{FALSE}$ when $\mu(s) = 0$, and $Q(s,a)$ is a convex combination of $\hat{y}_a^{TRUE}$ and $\hat{y}_a^{TRUE}$ otherwise. 
% The Q-learning update for each of the parameters of the tree are shown in Eq.~\ref{eq:QUpdate}.
% \par\nobreak{\parskip0pt \small 
% \begin{align}
% ^{Q}\Delta \theta &= \alpha \left(R(s,a) + \gamma\max_{a' \in A} Q^{\pi}_{\phi}(s',a') - Q^{\pi}_{\phi}(s,a) \right)\nabla Q^{\pi}_{\phi}(s,a) \\
% \rightarrow \alpha \left(R(s,a) + \gamma\max_{a' \in A} Q^{\pi}_{\phi}(s',a') - Q^{\pi}_{\phi}(s,a) \right)\nabla Q^{\pi}_{\phi}(s,a)
% \end{align}

% For example, if $\phi = 2$ and the agent's state index is 3 ($s=s_3$), the predicted Q-values for taking actions $a_1$ and $a_2$ are the values in the leaf node {\small $\hat{y}^{TRUE}$} ({\small$\hat{y}^{TRUE}_{a_1}$} and {\small$\hat{y}^{TRUE}_{a_2}$} for $a_1$ and $a_2$, respectively). 

% \subsection{Decision Trees as Function Approximators}
% \label{sec:DDTFA}
% We aim to learn the parameters of a DDT that can serve as either the function approximator for Q-learning or a policy for PG. 
\begin{figure*}
\begin{subfigure}{.45\linewidth}
  \centering
  \includegraphics[width=0.9\linewidth, height = 2.5cm]{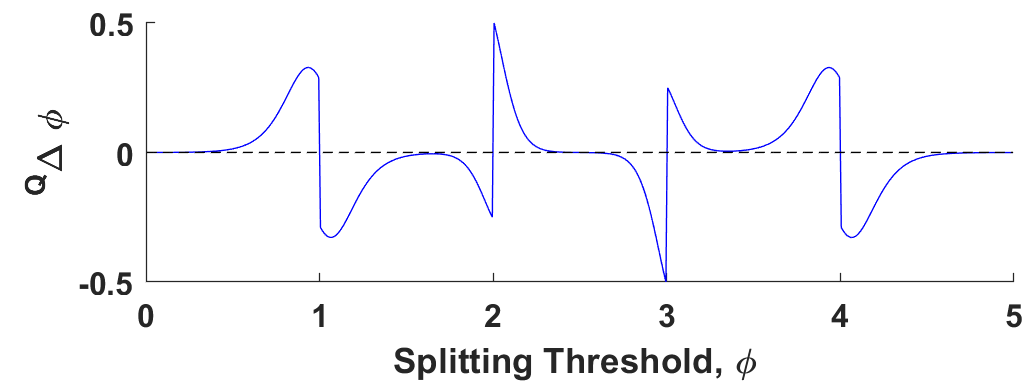}
  \caption{}
  \label{fig:CriticalPointsQL}
\end{subfigure}
\begin{subfigure}{0.55\linewidth}
  \centering
  \includegraphics[width=0.9\linewidth, height = 2.5cm]{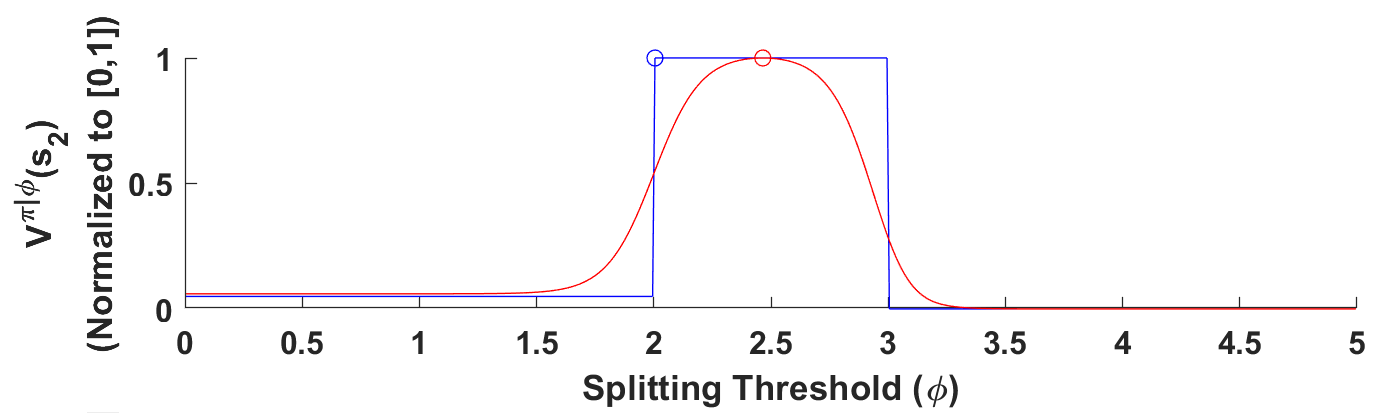}
  \caption{}
  \label{fig:PolicyValues}
\end{subfigure}
\begin{subfigure}{.45\linewidth}
  \centering
  \includegraphics[width=0.9\linewidth, height = 2.5cm]{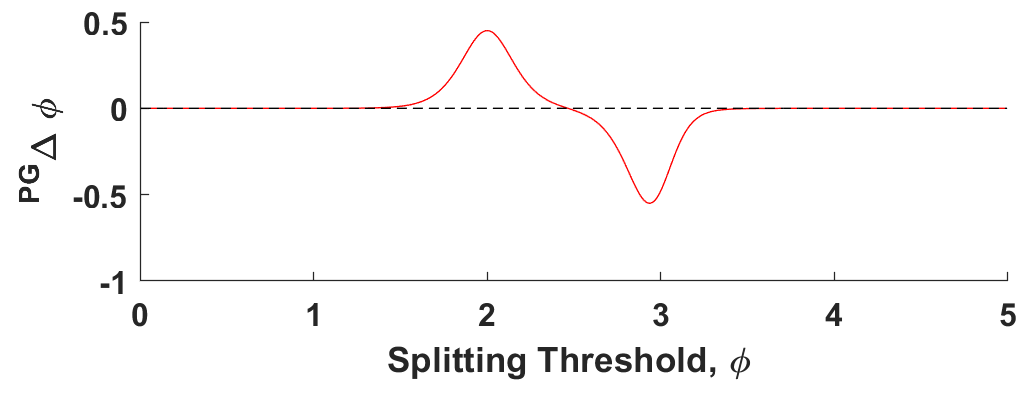}
  \caption{}
  \label{fig:CriticalPointsPG}
\end{subfigure}
\begin{subfigure}{0.55\linewidth}
  \centering
  \includegraphics[width=0.9\linewidth, height = 2.5cm]{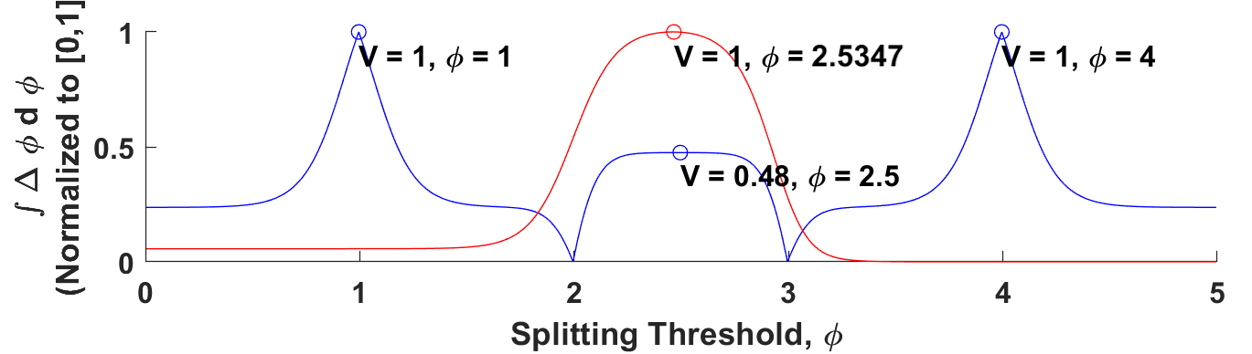}
  \caption{}
  \label{fig:OptimalityCurves}
\end{subfigure}
\caption{Figures~\ref{fig:CriticalPointsQL} and \ref{fig:CriticalPointsPG} depict the Q-learning and policy gradient update curves, respectively. Figures~\ref{fig:PolicyValues} and \ref{fig:OptimalityCurves} depict the policy value, $V^{\pi}$ (Fig.~\ref{fig:PolicyValues}), and the integrated gradient updates (Fig.~\ref{fig:OptimalityCurves}) for Q-learning (blue) and policy gradient (red) for the MDP depicted in Fig.~\ref{fig:MDP}.}
\label{fig:GradientUpdates}
\end{figure*}

\section{Interpretability for Online Learning}
\label{sec:interpretable-gradient}

We seek to address the two key drawbacks of the original DDT formulation by Su\'{a}rez and Lutsko \citep{suarez1999globally} in making the tree interpretable. First, the operation $\beta_\eta^\top x$ at each node produces a linear combination of the features, rather than a single feature comparison. Second, use of the sigmoid activation fuction means that there is a smooth transition between the $TRUE$ and $FALSE$ evaluations of a node, rather than a discrete decision. We address these limitations below; we demonstrate the extensibility of our approach by also differentiating over a rule list architecture \citep{letham2015interpretable} and extracting interpretable rule lists. Using the mechanisms from Sections \ref{subsec:discretizing} and \ref{subsec:discretizing-rules}, we produce interpretable policies for empirical evaluation (Section \ref{sec:results}) and a user study (Section \ref{sec:study}).

\subsection{Discretizing the Differentiable Decision Tree}
\label{subsec:discretizing}
Due to the nature of the sigmoid function, even a sparse $\beta_\eta$ is not sufficient to guarantee a discrete decision at each node. Thus, to obtain a truly discrete tree, we convert the differentiable tree into a discrete tree by employing an $\argmax_j(\beta_\eta^j)$ to obtain the index of the feature of $j$ that the node will use. We set $\beta_\eta$ to a one-hot vector, with a $1$ at index $j$ and $0$ elsewhere. We also divide $\phi_\eta$ by the node's weight $\beta_\eta^j$, normalizing the value for comparison against the raw input feature $x_j$. Each node then compares a single raw input feature to a single $\phi_\eta$, effectively converting from Eq.~\ref{eq:sigmoid-function} back into Eq.~\ref{eq:basicSplit}. We repeat this process for each decision node, obtaining discrete splits throughout the tree. Finally, each leaf node must now return a single action, as in an ordinary decision tree. We again employ an $\argmax_j(\beta_\eta^j)$ on each leaf node and set the leaves to be one-hot vectors with $\beta_\eta^j = 1$ and all other values set to $0$. The result of this process is an interpretable decision tree with discrete decision nodes, a single feature comparison per node, and a single decision output per leaf.

% The above process is repeatable with any alteration to the DDT architecture, including the rule-list extension. We employ this method for extracting interpretable decision lists in addition to standard binary trees.

\subsection{Differentiable Rule Lists}
\label{subsec:discretizing-rules}
In addition to discretizing the optimized tree parameterization, we also consider a specific sub-formulation of tree proposed by \citep{letham2015interpretable} to be particularly interpretable: the rule- or decision-list. This type of tree restricts the symmetric branching allowed for in Eq.~\ref{eq:yhat} by stating that the TRUE branch from a decision node leads directly to a leaf node. We define a discrete rule list according to Eq.~\ref{eq:ruleList}.
\par\nobreak{\parskip0pt \footnotesize \begin{align}
T_{\eta}(x) &\coloneqq  
\begin{cases} 
  y_{\eta}, &  \text{if leaf} \\ % \nexists\eta_\swarrow \\
  \mu_{\eta}(x)y_{\eta_\swarrow} + (1-\mu_{\eta}(x))T_{\eta_\searrow}(x) & \text{o/w}
\end{cases}\label{eq:ruleList}
\end{align}}In Section \ref{sec:results}, we demonstrate that these mechanisms for interpretability achieves high-quality policies for online RL and are consistent with the the legal~\citep{voigt2017eu} and practical criteria for interpretability \citep{doshi2017towards,letham2015interpretable}.

\begin{figure*}
\centering
\begin{subfigure}{0.475\linewidth}
  \centering
  \includegraphics[width=.8\linewidth, height = 3.25cm]{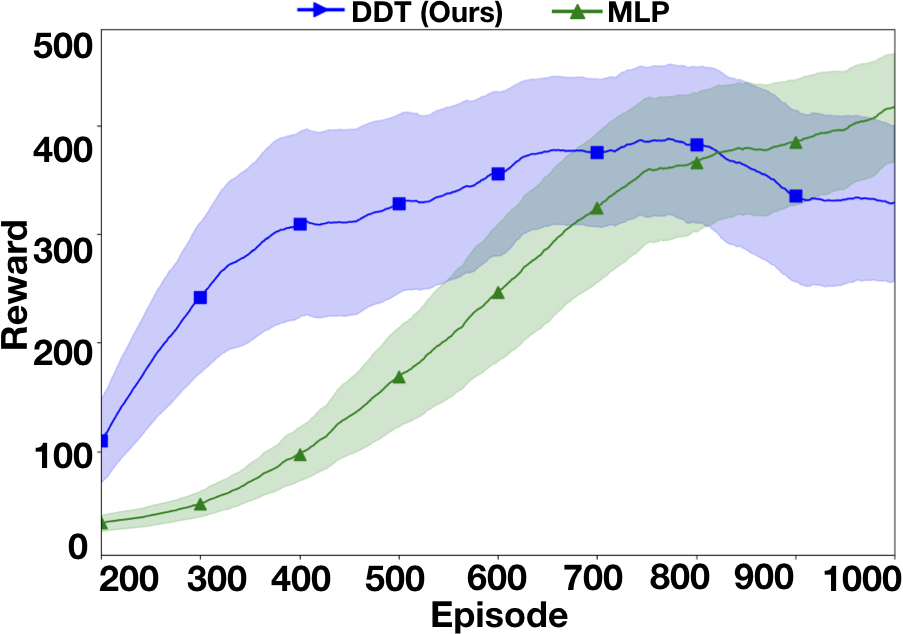}
\end{subfigure}
\begin{subfigure}{0.475\linewidth}
  \centering
    \includegraphics[width=.8\linewidth, height = 3.25cm]{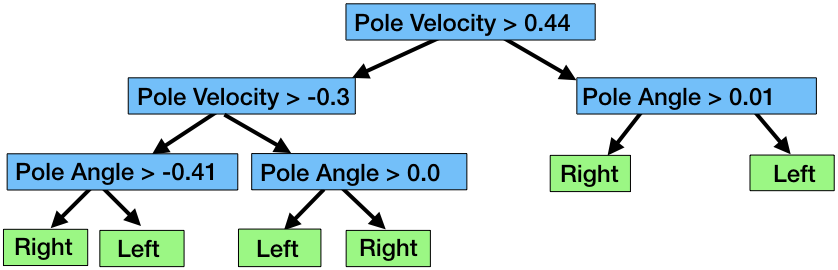}
\end{subfigure}
\caption{Training curves for the cart pole domain (left), and the resulting discrete decision tree (right)}
\label{fig:cart-results}
\vspace{-0.3cm}
\end{figure*}

\section{Analysis of Gradient Methods for DDTs}
\label{sec:analytical}
In this section, we analyze Q-learning and policy gradient updates for DDTs as function approximators in RL, providing a theoretical basis for how to best deploy DDTs to RL. We show that Q-learning introduces additional critical points that impede learning where policy gradient does not. This analysis guides us to recommend policy gradient for these interpretable function approximators and yields high-quality policies (Section \ref{sec:results}).

\subsection{Analysis: Problem Setup} 
We consider an MDP with states $S = \{s_1,s_2,...,s_n\}$ and actions $A = \{a_1,a_2\}$ (Figure ~\ref{fig:MDP}). The agent moves to a state with a higher index (i.e., $s = s+1$) when taking action $a_1$ with probability $p$ and $1-p$ for transitioning to a lower index. The opposite is the case for action $a_2$. Within Figure ~\ref{fig:MDP}, $a_1$ corresponds to ``move right'' and $a_2$ corresponds to ``move left.'' The terminal states are $s_1$ and $s_n$. {{The rewards are zero for each state except for $R(s_{i^*})=R(s_{i^*+1})=+1$ for some $i^*$ such that $1 < i^* < n-1$. % This reward schema encourages the agent to apply $a_1$ in state $s_i^*$ and $a_2$ in state $s_i^*$ to maximize its reward. 
It follows that the optimal policy, $\pi^*$, is $\pi(s)=a_1$ (``move right'') in $s_j$ such that $1 \leq j \leq i^*$ and $\pi(s)=a_2$ otherwise.}} A proof is given in supplementary material. We optimistically assume $p = 1$; despite this hopeful assumption, we show unfavorable results for Q-learning and policy-gradient-based agents using DDTs as function approximators. 

\subsection{Analysis: Tree Initialization}
For our investigation, we assume that the decision tree's parameters are initialized to the optimal setting. Given our MDP setup, we only have one state feature: the state's index. As such, we only have two degrees of freedom in the decision node: the steepness parameter, $\alpha$, and the splitting criterion, $\phi$. In our analysis, we focus on this splitting criterion, $\phi$, showing that even for an optimal tree initialization, $\phi$ is not compelled to converge to the optimal setting $\phi = i^* + \frac{1}{2}$. We set the leaf nodes as follows for Q-learning and policy gradient given the optimal policy. 

For Q-learning, we set the discounted optimal action reward as {$\hat{q}^{TRUE}_{a_2}=\hat{q}^{FALSE}_{a_1}=\sum_{t=0}^\infty{\gamma^t r^+ = \frac{r^+}{1-\gamma}}$}, which assumes $s_o=s_{i^*}$. Likewise, we set $\hat{y}^{TRUE}_{a_1}=\hat{y}^{FALSE}_{a_2} = \frac{r^+}{1-\gamma} - r^+ - r^+\gamma$, which correspond to the Q-values of taking action $a_1$ and $a_2$ in states $s_2$ and $s_3$ when otherwise following the optimal policy starting in a non-terminal node.

For policy gradient, we set {\small$\hat{y}^{TRUE}_{a_2}=\hat{y}^{FALSE}_{a_1}=0.99$} and {\small$\hat{y}^{TRUE}_{a_1}=\hat{y}^{FALSE}_{a_2} = 0.01$}. These settings correspond to a decision tree that focuses on exploiting the current (optimal if $\phi=i^*$) policy. While we consider this setting of parameters for our analysis of DDTs, the results generalize.

\begin{figure*}
\centering
\begin{subfigure}{0.475\linewidth}
  \centering
  \includegraphics[width=.8\linewidth, height = 3.25cm]{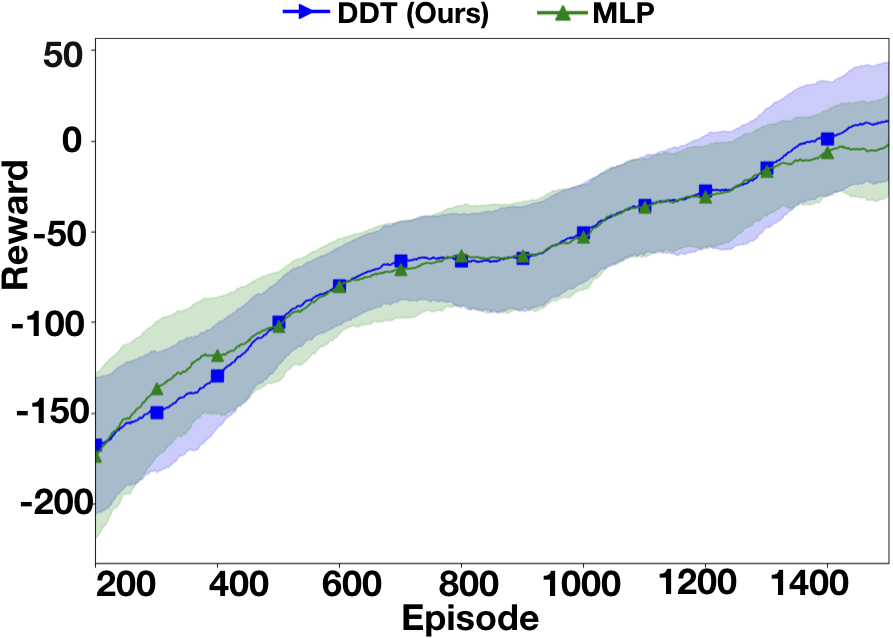}
\end{subfigure}
\begin{subfigure}{0.475\linewidth}
  \centering
    \includegraphics[width=.8\linewidth, height = 3.25cm]{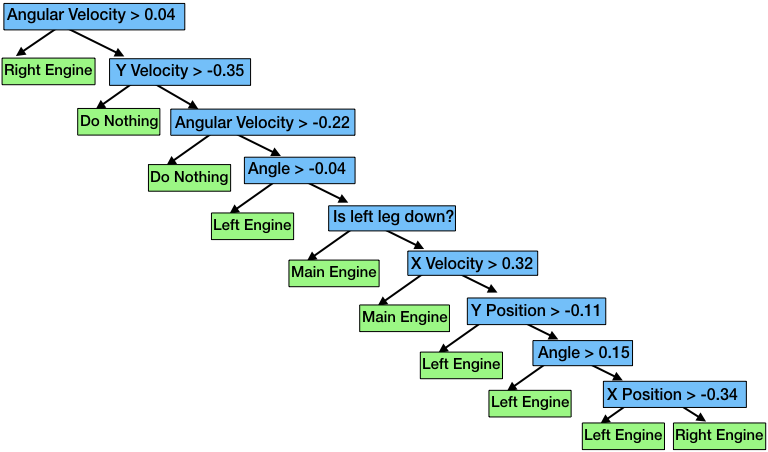}
\end{subfigure}
\caption{Training curves for the lunar lander domain (left), and the resulting discrete rule list (right)}
\label{fig:lunar-results}
\vspace{-0.3cm}
\end{figure*}

\subsection{Computing Critical Points}
\label{sec:ComputingCRs}

The ultimate step in our analysis is to assess whether Q-learning or policy gradient introduces critical points that do not coincide with global extrema. To do so, we can set Equations \ref{eq:QLearningPolicy} and \ref{eq:policyGradient} to zero, with $\nabla Q(s,a) = \nabla f_T{s,a}$ from Eq.~\ref{eq:QTree} and $\nabla \pi(s,a) = \nabla f_T(s,a)$ from Eq.~\ref{eq:PGTree}, respectively. We would then solve for our parameter(s) of interest and determine whether any zeros lie at local extrema. In our case, focusing on the splitting criterion, $\phi$, is sufficient to show the weaknesses of Q-learning for DDTs. 

Rather than exactly solving for the zeros, we use numerical approximation for these Monte Carlo updates (Equations \ref{eq:QLearningPolicy} and \ref{eq:policyGradient}). In this setting, we recall that the agent experiences episodes with $T$ timesteps. Each step generates its own update, which are combined to give the overall update $\Delta \phi = \sum_{t=0}^T \Delta \phi^{(t)}$. Pseudo-critical points exist, then, whenever $\Delta \phi$ = 0. A gradient descent algorithm would treat these as extrema, and the gradient update would push $\phi$ towards these points. As such, we consider these ``critical points.''

\subsection{Numerical Analysis of the Updates}
The critical points given by $\Delta \phi = 0$ are shown in Fig.~\ref{fig:CriticalPointsQL} and \ref{fig:CriticalPointsPG} for Q-learning and PG, respectively. For the purpose of illustration, we set $n = 4$ (i.e., the MDP has four states). As such, $i^* = 2$ and the optimal setting for $\phi=\phi^*=i^* + \frac{1}{2} = 2.5$.

For Q-learning, there are five critical points, only one of which is coincident with $\phi = \phi^* = i^* + \frac{1}{2}$. For PG, there are fewer, with a single critical point in the domain of $\phi\in(-\infty,\infty)$, which occurs at $\phi \approx 2.465$\footnote{We recall that, for this analysis, $s_o = i^*$; if we set $s_o = i^*+1$ (i.e., the symmetric position with respect to vertical), this critical point for policy gradient is $\phi = 2.535$.}.  Thus, we can say that the expectation of the critical point for a random, symmetric initialization is $\E_{s_o \sim U(2,3)}[\Delta \phi = 0 | s_o] =  i^* + \frac{1}{2}$, which supports the adoption of policy gradient as an approach for DDTs.

Additionally, by integrating $\Delta \phi$ with respect to $\phi$ from $0$ to $\phi$, i.e., $\text{Optimality}(\phi) = \int_{\phi'=0}^{\phi} \Delta \phi' d\phi'$, we infer the ``optimality curve,'' which should equal the value of the policy, $V^{\pi_\phi}$, implied by Q-learning and policy gradient. We numerically integrate using Riemann's method normalized to $[0,1]$. One would expect that the respective curves for the policy value (Figure \ref{fig:PolicyValues}) and integrated gradient updates (Figure \ref{fig:OptimalityCurves}) would be identical; however, this does not hold for Q-learning. Q-learning with DDT function approximation introduces undesired extrema, shown by the blue curve in Figure \ref{fig:OptimalityCurves}. Policy gradient, on the other hand, maintains a single maximum coincident with $\phi = \phi^* = i^* + \frac{1}{2} = 2.5$. 

This analysis provides evidence that Q-learning exhibits weaknesses when applied to DDT models, such as an excess of critical points which serve to impede gradient descent. We therefore conclude that policy gradient is a more promising approach for learning the parameters of DDTs and proceed accordingly. As such, we have {{shown}} that Q-learning with DDT function approximators introduces additional extrema that policy gradients, under the same conditions, do not, within our MDP case study.
%\red{Theorem \ref{thm:Main}.}
%\red{\begin{theorem}[Q-Learning introduces additional extrema]
% Q-learning with DDT function approximation introduces additional extrema that policy gradients under the same conditions do not for our MDP case study.
% \label{thm:Main}
% \end{theorem}}

This analysis provides the first examination of the potential pitfalls and failings of Q-learning with DDTs. We believe that this helpful analysis will guide researchers in the application of these function approximators. Given this analysis and our mechanisms for interpretability (Section \ref{sec:interpretable-gradient}), we now show convincing empirical results (Section \ref{sec:results}) of the power of these function approximators to achieve high-quality and interpretable policies in RL.

\begin{figure*}
\begin{subfigure}{0.45 \linewidth}
  \centering
  \includegraphics[width=.8\linewidth, height = 3.25cm]{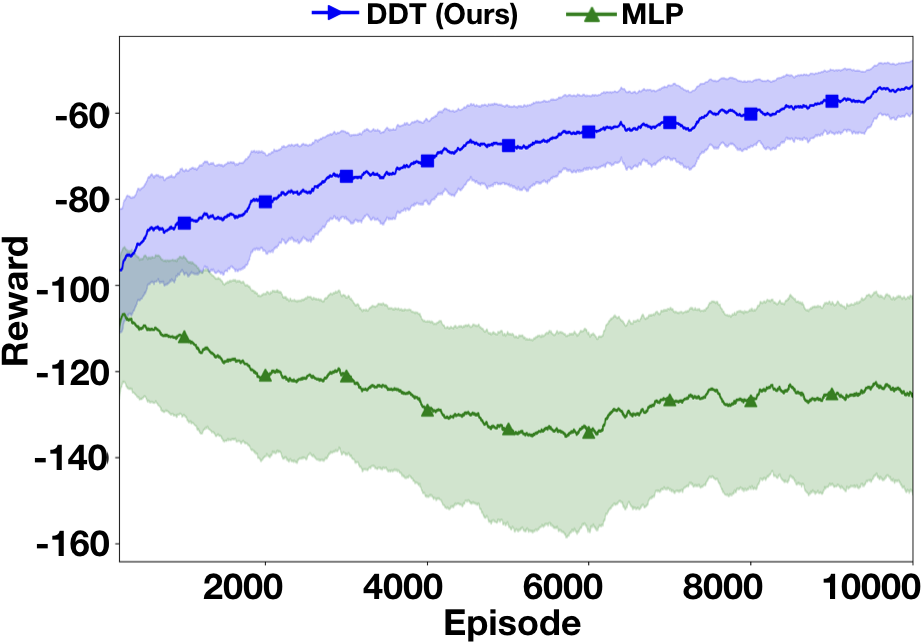}
\end{subfigure}
\begin{subfigure}{0.55 \linewidth}
  \centering
    \includegraphics[width=.8\linewidth, height = 3.25cm]{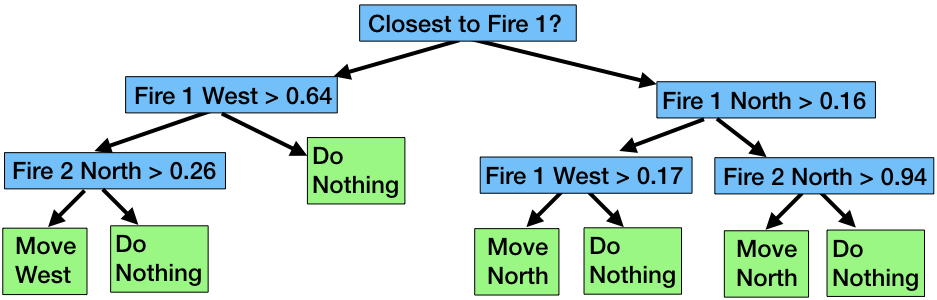}
\end{subfigure}
\caption{Training curves for the wildfire tracking environment (left) and the resulting discrete decision tree (right)}
\label{fig:fire-results}
\vspace{-0.1cm}
\end{figure*}

\section{Demonstration of DDTs for Online RL}
\label{sec:results}
Our ultimate goal is to show that DDTs can learn competent, interpretable policies online for RL tasks. To demonstrate this, we evaluate our DDT algorithm using the cart pole and lunar lander OpenAI Gym environments \citep{gym:2016}, a simulated wildfire tracking problem, and the FindAndDefeatZerglings mini-game from the StarCraft II Learning Environment \citep{vinyals2017starcraft}. All agents are trained via Proximal Policy Optimization (PPO) \citep{ pytorchrl,schulman2017proximal}.  We use a multilayer perceptron (MLP) architecture as a baseline for performance across all tasks. We provide further details on the evaluation domains below, as well as examples of extracted interpretable policies, trained using online RL with DDTs. Due to space constraints, we present pruned versions of the interpretable policies in which redundant nodes are removed for visual clarity. The full policies are in the supplementary material.

We conduct a sensitivity analysis comparing the performance of MLPs with DDTs (DDTs) across a range of depths. For the trees, the set of leaf nodes we consider is \{2, 4, 8, 16, 32\}. For comparison, we run MLP agents with between \{0, 1, 2, 4, 8, 16, 32\} hidden layers, and a rule-list architecture with \{1, 2, 4, 8, 16, 32\} rules. Results from this sensitivity analysis are given in Fig.~4 \& 5 in the supplementary material. We find that MLPs succeed only with a narrow subset of architectures, while DDTs and rule lists are more robust. In this section, we present results from the agents that obtained the highest average cumulative reward in our sensitivity analysis. Table \ref{tab:results} compares mean reward of the highest-achieving agents and shows the mean reward for our discretization approach applied to the best agents. For completeness, we also compare against standard decision trees which are fit using scikit-learn \citep{scikit-learn} on a set of state-action pairs generated by the best-performing model in each domain, which we call \textit{State-Action DT}. 

In our OpenAI Gym \citep{gym:2016} environments we use a learning rate of 1e-2, and in our wildfire tracking and FindAndDefeatZerglings \citep{vinyals2017starcraft} domains we use a learning rate of 1e-3. All models are updated with the RMSProp \citep{tieleman2012lecture} optimizer. All hyperparameters are included in the supplementary material.

\begin{table*}[t]
\caption{Average cumulative reward for top models across methods and domains. Bold denotes highest-performing method.} \label{tab:results}
\begin{center}
\begin{tabular}{lcccc}
\textbf{Agent Type} &\textbf{Cart Pole} &\textbf{Lunar Lander} &\textbf{Wildfire Tracking} &\textbf{FindAndDefeatZerglings} \\
\hline\hline
DDT Balanced Tree (ours) &  \textbf{500 $\pm$ 0} & \textbf{97.9 $\pm$ 10.5} & \textbf{-32.0 $\pm$ 3.8}  & 6.6 $\pm$ 1.1\\
DDT Rule List (ours) &  \textbf{500 $\pm$ 0} & 84.5 $\pm$ 13.6 &  -32.3 $\pm$ 4.8  & {\textbf{11.3 $\pm$ 1.4}}\\
MLP & \textbf{500 $\pm$ 0}  & 87.7 $\pm$ 21.3 & -86.7 $\pm$ 9.0 & {6.6 $\pm$ 1.2} \\ \hline
Discretized DDT (ours) & \textbf{499.5 $\pm$ 0.8} & -88 $\pm$ 20.4 & \textbf{-36.4 $\pm$ 2.6} & \textbf{4.2 $\pm$ 1.6} \\
Discretized Rule List (ours) & 414.4 $\pm$ 63.9 & \textbf{-78.4  $\pm$ 32.2} & -39.8 $\pm$ 1.8 & 0.7 $\pm$ 1.3 \\
State-Action DT & 66.3 $\pm$ 18.5 & -280.9 $\pm$ 60.6 & -67.9 $\pm$ 7.9 & -3.0 $\pm$ 0.0 \\
\end{tabular}
\end{center}
\vspace{-0.3cm}
\end{table*}
\vspace{-0.1cm}
\subsection{Open AI Gym Evaluation}
\label{subsec:open-ai-results}

We plot the performance of the best agent for each architecture in our OpenAI Gym \citep{gym:2016} experiments, as well as pruned interpretable policies, in Fig.~\ref{fig:cart-results} and \ref{fig:lunar-results}. To show the variance of the policies, we run five seeds for each policy-environment combination. Given the flexibility of MLPs and their large number of parameters, we anticipate an advantage in raw performance. We find that the DDT offers competitive or even superior performance compared to the MLP baseline, and even after converting the trained DDT into a discretized, interpretable tree, the training process yields tree policies that are competitive with the best MLP. {{Our interpretable approach yields a 3x and 7x improvement over a batch-trained decision tree (DT) on lunar lander and cart pole, respectively.}} Table \ref{tab:results} depicts the average reward across domains and agents.

\subsection{Wildfire Tracking}
\label{subsec:wildfire-results}
Wildfire tracking is a real-world problem in which interpretability is critical to an agent's human teammates. While RL is a promising approach to develop assistive agents in wildfire monitoring~\citep{haksar2018distributed}, it is important to maintain trust between these agents and humans in this dangerous domain. An agent that can explicitly give its policy to a human firefighter is therefore highly desirable.

We develop a Python implementation of the simplified FARSITE \citep{finney1998farsite} wildfire propagation model. The environment is a 500x500 map in which two fires propagate slowly from the southeast end of the map to the northwest end of the map and two drones are randomly instantiated in the map. Each drone receives a 6D state containing distances to each fire centroid and Boolean flags indicating which fire the drone is closer to. The RL agent is duplicated at the start of each episode and applied to each drone, and the drones do not have any way of communicating. The challenge is to identify which fire is closest to the drone, and to then take action to get as close as possible to that fire centroid, with the objective of flying above the fires as they progress across the map. Available actions include four move commands (north, east, south, west) and a ``do nothing'' command. The reward function is the negative distance from drones to fires, given in Eq.~\ref{eqn:fire-reward} where $D$ is a distance function, $d_i$ are the drones, and $f_i$ are the fires.
\vspace{-0.5cm}
\par\nobreak{\parskip0pt \small
\begin{align}
 R &= - \min \left[(D(d_1, f_1), D(d2, f1))\right] \nonumber \\
  &- \min \left[D(d1, f2), D(d2, f2)\right] \label{eqn:fire-reward}
\end{align}}The reward over time for the top performing DDT and MLP agents is given in Figure \ref{fig:fire-results}, showing the DDT significantly outperforms the MLP. We also present the interpretable policy for the best DDT agent, which has the agent neglect the south and east actions, instead checking for north and west distances and moving in those directions. This behavior reflects the dynamics of the domain, in which the fire always spreads from southeast to northwest. {{The best interpretable policy we learn is $\approx$2x better than the best batch-learned tree and $>$2x better than the best MLP.}}

\begin{figure*}
\begin{subfigure}{0.45 \linewidth}
  \centering
  \includegraphics[width=.8\linewidth, height = 3.25cm]{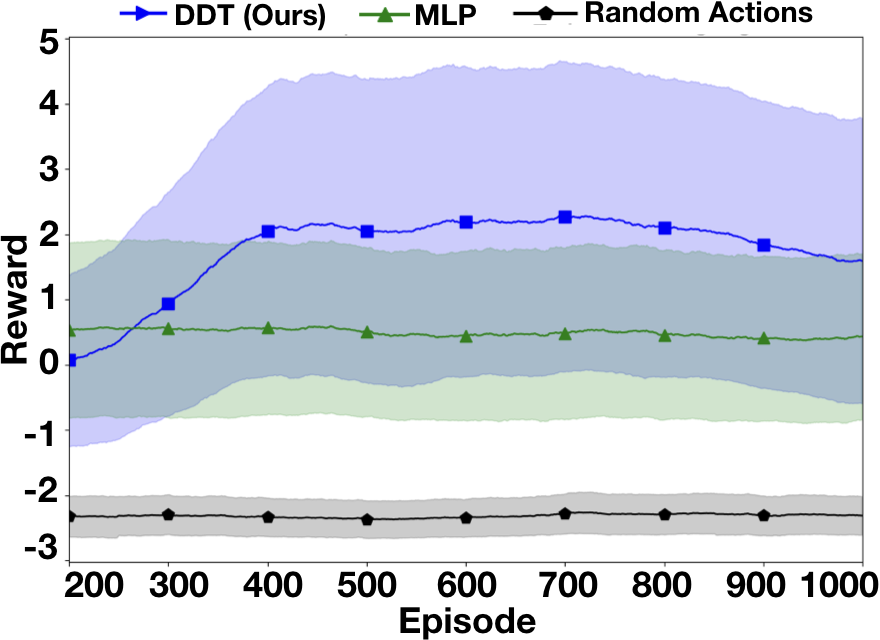}
\end{subfigure}
\begin{subfigure}{0.55 \linewidth}
  \centering
    \includegraphics[width=.8\linewidth, height = 3.25cm]{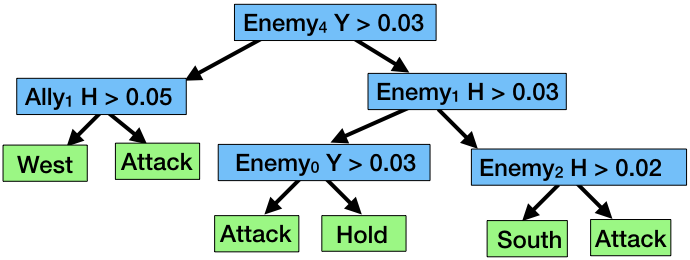}
\end{subfigure}
\caption{Training curves for the FindAndDefeatZerglings environment (left) and the resulting discrete decision tree (right)}
\label{fig:zerg-results}
\vspace{-0.3cm}
\end{figure*}

\vspace{-0.1cm}
\subsection{StarCraft II Micro-battle Evaluation}
\label{subsec:sc2-results}

To further evaluate the DDT and discretized tree, we use the FindAndDefeatZerglings minigame from the StarCraft II Learning Environment \citep{vinyals2017starcraft}. For this challenge, three allied units explore a partially observable map and defeat as many enemy units as possible within three minutes. We assign each allied unit a copy of the same learning agent. Rather than considering the image input and keyboard and mouse output, we manufacture a reduced state-action space. The input state is a 37D vector of allied and visible enemy state information, and the action space is 10D consisting of move and attack commands. More information is in supplementary material.

As we can see in Figure \ref{fig:zerg-results}, our DDT agent is again competitive with the MLP agent {and is $>$2x better than a batch-learned decision tree.} The interpretable policy for the best DDT agent reveals that the agent frequently chooses to attack, and never moves in conflicting directions. This behavior is intuitive, as the three allied units should stay grouped to increase their chances of survival. The agent has learned not to send units in conflicting directions, instead moving units southwest while attacking enemies en route.
\vspace{-0.3cm}
\section{Interpretability Study}
\label{sec:study}
\vspace{-0.2cm}
To emphasize the interpretability afforded by our approach, we conducted a user study in which participants were presented with policies trained in the cart pole domain and tasked with identifying which decisions the policies would have made given a set of state inputs. We compared interpretability between a discrete decision tree, a decision list, and a one-hot MLP without activation functions.

\subsection{Study Setup}
\label{subsec:study-setup}
We designed an online questionnaire to survey 15 participants, giving each a discretized DDT, a discretized decision list, and a sample one-hot MLP. The discretized policies are actual policies from our experiments, presented in Table \ref{tab:results}. Rather than include the full MLP, which is available in the supplementary material, we binarized the weights, thereby make the calculation much easier and less frustrating for participants. This mechanism is similar to current approaches to interpretability with deep networks that use attention \citep{serrano2019attention} so that human operators can see what the agent is considering when it makes a decision.

After being given a policy, participants were presented with five sample domain states. They were then asked to trace the policy with a given input state and predict what the agent would have done. After predicting which decisions the agent would have made, participants were presented with a set of Likert scales assessing their feelings on the interpretability and usability of the given policy as a decision-making aid. We timed participants for each method.

We hypothesize that: \textbf{H1:} \emph{A decision-based classifier is more interpretable than an MLP}; \textbf{H2:} \emph{A decision-based classifier is more efficient than a MLP.} To test these hypotheses, we report on participant Likert scale ratings (\textbf{H1}) and completion time for each task (\textbf{H2}).
\vspace{-0.3cm}
\begin{figure}[b]
    \centering
    \includegraphics[width=0.9\linewidth, height = 4.5cm]{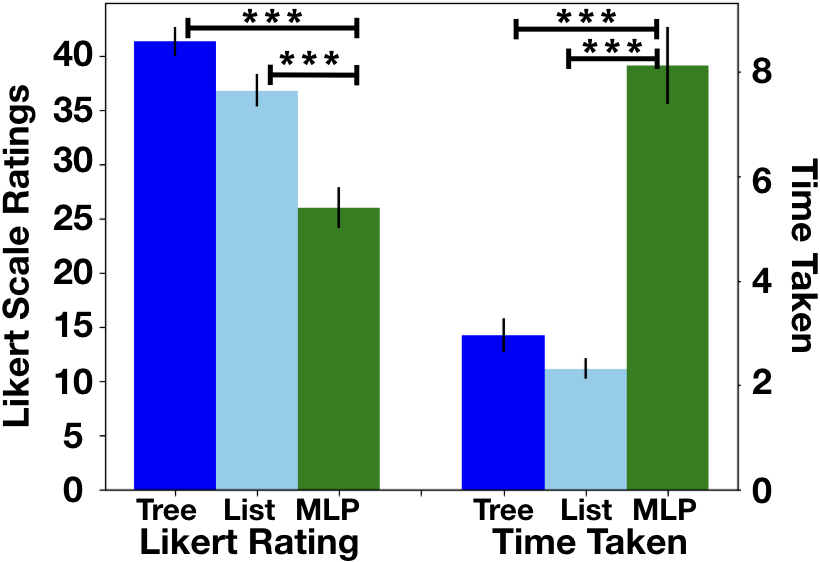}
    \caption{Results from our user study. Higher Likert ratings are better, lower time taken is better.}
    \label{fig:study-results}
\end{figure}

\subsection{Study Results}
\label{subsec:study-results}
Results of our study are shown in Figure \ref{fig:study-results}. We perform an ANOVA and find that the type of decision-making aid had a statistically significant effect on users' Likert scale ratings for usability and interpretability ($F(2, 28) = 19.12, p < 0.0001$). We test for normality and homoscedasticity and do not reject the null hypothesis in either case, using Shapiro-Wilk ($p > 0.20$) and Levene's Test ($p > 0.40$), respectively. A Tukey's HSD post-hoc test shows that the tree ($t = 6.02$, $p < 0.0001$) and decision list ($t = 4.24$, $p < 0.0001$) both rated significantly higher than a one-hot MLP.

We also test the time participants took to use each decision-making aid for a set of five prompts. We applied Friedman's test and found the type of aid had a significant effect on completion time ($Q(2) = 26, p < 0.0001$). Dunn's test showed that the tree ($z = -4.07$, $p < 0.0001$) and decision list ($z = -5.23$, $p < 0.0001$) times were statistically significantly shorter than the one-hot MLP completion times.

We note that participants were shown the full MLP after the questionnaire's conclusion, and participants consistently reported they would have abandoned the task if they had been presented with a full MLP as their aid. These results support the hypothesis that decision trees and lists are significantly superior decision-making aids in reducing human frustration and increasing efficiency. This study, coupled with our strong performance results over MLPs, shows the power of our approach to interpretable, online RL via DDTs. 

\vspace{-0.3cm}

\section{Future Work}
 \label{sec:futureWork}
 \vspace{-0.3cm}

We propose investigating how our framework could leverage advances in other areas of deep learning, e.g. inferring feature embeddings. For example, we could learn subject-specific embeddings via backpropagation but within an interpretable framework for personalized medicine~\citep{killian2017robust} or in apprenticeship learning~\citep{gombolay2018robotic}, particularly when heterogeneity precludes a one-size-fits-all model~\citep{chen2020joint}. We could also invert our learning process to a prior specify a decision tree policy given expert knowledge, which we could then train via policy gradient~\citep{silva2019prolonets}.
 \vspace{-0.3cm}

\section{Conclusion}
 \label{sec:conclusion}
 \vspace{-0.3cm}

We demonstrate that DDTs can be used in RL to generate interpretable policies. We provide a motivating analysis showing the benefit of using policy gradients to train DDTs in RL challenges over Q-learning. This analysis serves to guide researchers and practitioners alike in future research and application of DDTs to learn interpretable RL policies. We show that DDTs trained with policy gradient can provide comparable and even superior performance against MLP baselines. Finally, we conduct a user study which demonstrates that DDTs and decision lists offer increased interpretability and usability over MLPs while also taking less time and providing efficient insight into policy behavior. 
% This research opens the door for further investigations in developing interpretable, online models for RL amenable to gradient descent.
\vspace{-0.3cm}

\section{Acknowledgements}
\vspace{-0.3cm}

This work was funded by the Office of Naval Research under grant N00014-19-1-2076 and by MIT Lincoln Laboratory under grant \# MIT Lincoln Laboratory grant
7000437192.

\bibliographystyle{plainnat}
\bibliography{paper}
\clearpage
\appendix
\section*{Appendix A: Derivation of the Optimal Policy}
% the \\ insures the section title is centered below the phrase: AppendixA
In this section, we provide a derivation of the optimal policy for the MDP in Fig.~\ref{fig:MDP}. For this derivation, we use the definition of the Q-function described in Eq.~\ref{eq:App0}, where $s'$ is the state resulting from applying action $a$ in state $s$. In keeping with the investigation in this paper, we assume deterministic transitions between states (i.e., $p=1$ from Eq.~\ref{eq:Transition}). As such, we can ignore $P(s'|s,a)$ and simply apply Eq.~\ref{eq:App00}.

\begin{equation}
P(s'|s,a) = 
    \begin{cases} 
    0, & \text{if } s' \in {1,4} \\
    p, & \text{if } (s' = s + 1, a = a_1)  \\
     &\indent\indent\indent  \lor  (s' = s - 1, a = a_2)\\
    \frac{1-p}{|S|-1}, &  \text{otherwise}
    \end{cases} 
\label{eq:Transition}
\end{equation}

\par\nobreak{\parskip0pt \small \noindent
\begin{align}
Q(s,a) &:= R(s,a) + \gamma \max_{a'} \sum_{s'} P(s'|s,a)Q(s',a') \label{eq:App0} \\
Q(s,a) &:= R(s,a) + \gamma \max_{a'} Q(s',a') \label{eq:App00}
\end{align}}
\begin{figure}[h]
  \centering
    \includegraphics[height = 3cm]{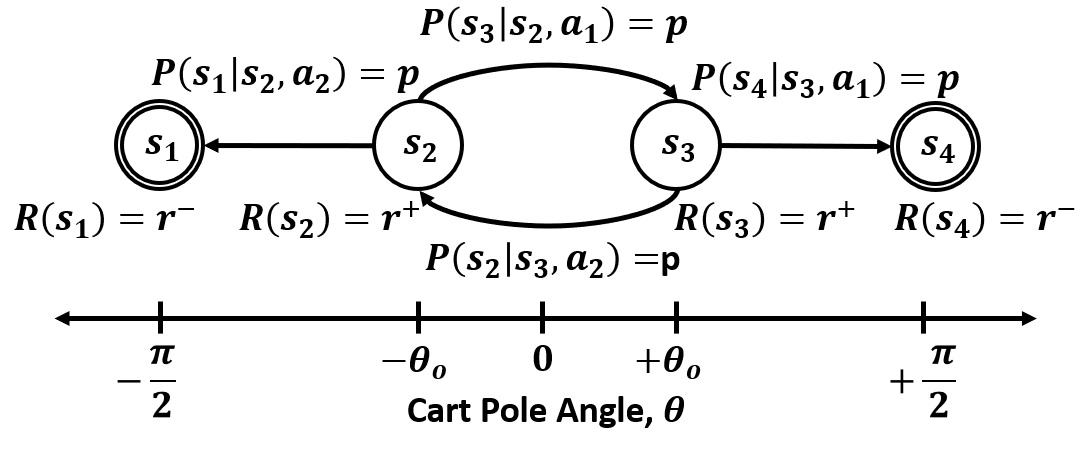}
    \caption{}
    \label{fig:MDP_appendix}\end{figure}
\begin{theorem}
The optimal policy for the MDP in Fig.~\ref{fig:MDP_appendix} is to apply action $a_1$ in state $s_2$ and action $a_2$ in state $s_3$ assuming deterministic transitions between states (i.e., $p=1$ from Eq.~\ref{eq:Transition}).
\label{thm:OptPol}
\end{theorem}

We begin by asserting in Eq.~\ref{eq:App1} that the Q-values for $Q(s,a)$ are $r^-$ given $s\in\{1,4\}$ and for any action $a \in \{a_1,a_2\}$. This result is due to the definition that states $s_1$ and $s_4$ are terminal states and the reward for those states is $r^-$ regardless of the action applied. We note that, in our example, $r^-=0$, but we leave it here for the sake of generality.
\begin{align}
Q(s_1,a_1) = Q(s_1,a_2) = Q(s_4,a_1) = Q(s_4,a_2) = r^- \label{eq:App1}
\end{align}Next, we must compute the Q-values for the remaining state-action pairs, as shown in Eq.~\ref{eq:App2}-\ref{eq:App3}.
\begin{align}
Q(s_2,a_1) &= R(s_2,a_1) + \gamma \max \{Q(s_3,a_1),Q(s_3,a_2)\} \label{eq:App2} \\
Q(s_2,a_2) &= R(s_2,a_1) + \gamma \max \{Q(s_1,a_1),Q(s_1,a_2)\} \\
Q(s_3,a_1) &= R(s_3,a_1) + \gamma \max \{Q(s_4,a_1),Q(s_4,a_2)\} \\
Q(s_3,a_2) &= R(s_3,a_2) + \gamma \max \{Q(s_2,a_1),Q(s_2,a_2)\} \label{eq:App3}
\end{align}
By the definition of the MDP in Fig.~\ref{fig:MDP_appendix}, we substitute in for $R(s_2,a_1)=R(s_2,a_2)=R(s_3,a_1)=R(s_3,a_2) = r^+$ as shown in Eq.~\ref{eq:App4}-\ref{eq:App5}.

\begin{align}
Q(s_2,a_1) &=  r^+        + \gamma \max \{Q(s_3,a_1),Q(s_3,a_2)\} \label{eq:App4}\\
Q(s_2,a_2) &=  r^+        + \gamma r^- \label{eq:App4b}\\
Q(s_3,a_1) &= r^+        + \gamma r^- \label{eq:App4c}\\
Q(s_3,a_2) &= r^+        + \gamma \max \{Q(s_2,a_1),Q(s_2,a_2)\} \label{eq:App5} 
\end{align}We can substitute in for $Q(s_3,a_1)$ and $Q(s_2,a_2)$ given Eq.~\ref{eq:App4} and \ref{eq:App5}.

\begin{align}
Q(s_2,a_1) &=  r^+        + \gamma \max \{\left(r^+ + \gamma r^-\right),Q(s_3,a_2)\} \label{eq:App6}\\
Q(s_3,a_2) &= r^+        + \gamma \max \{Q(s_2,a_1),\left(r^+ + \gamma r^-\right)\} \label{eq:App7} 
\end{align}

For the Q-value of state-action pair, $Q(s_2,a_1)$, we must determine whether $\left(r^+ + \gamma r^-\right)$ is less than or equal to $Q(s_3,a_2)$. If the agent were to apply action $a_2$ in state $s_3$, we can see from Eq.~\ref{eq:App7} that the agent would receive at a minimum $Q(s_3,a_2)\geq r^+ + \gamma\left(r^+ + \gamma r^-\right)$, because  $r^+ + \gamma\left(r^+ + \gamma r^-\right) > r^+ + \gamma r^-$, $Q(s_3,a_2)$ must be the maximum from Eq.~\ref{eq:App6}. We can make a symmetric argument for $Q(s_3,a_2)$ in Eq.~\ref{eq:App7}. Given this relation, we arrive at Eq.~\ref{eq:App8} and \ref{eq:App9}.
\begin{align}
Q(s_2,a_1) &=  r^+        + \gamma Q(s_3,a_2) \label{eq:App8}\\
Q(s_3,a_2) &= r^+        + \gamma Q(s_2,a_1) \label{eq:App9} 
\end{align}Eq.~\ref{eq:App8} and \ref{eq:App9} represent a recursive, infinite geometric series, as depicted in Eq.~\ref{eq:AppRec}.
\begin{align}
Q(s_2,a_1) = Q(s_3,a_2)  &=  r^+        + \gamma r^+ + \gamma^2 r^+ +\ldots \nonumber\\
         &=  r^+\left(\gamma^0 + \gamma + \gamma^2  +\ldots \right) \label{eq:App10} \\
  &= r^+ \sum_{t=0}^T \gamma^t \label{eq:AppRec} 
\end{align}In the case that $T=\infty$, Eq.~\ref{eq:AppRec} represents an infinite geometric series, the solution to which is $ \frac{r^+}{1+\gamma}$. In our case however, $T=3$ (i.e., four-time steps). As such, $Q(s_2,a_1) = Q(s_3,a_2) = r^+(1+\gamma +\gamma^2 + \gamma^3)$, as shown in Eq.~\ref{eq:Rap}.
\begin{align}
Q(s_2,a_1) = Q(s_3,a_2) = r^+(1+\gamma +\gamma^2 + \gamma^3)
\label{eq:Rap}
\end{align}

Recall that $r^- < 0$ given our definition of the MDP in Fig.~\ref{fig:MDP_appendix}. Therefore, $Q(s_2,a_1)=Q(s_3,a_2) = \frac{r^+}{1-\gamma} \geq Q(s_2,a_2)=Q(s_3,a_1) = r^+ + \gamma r^-$.
If the RL agent is non-myopic, i.e., $\gamma \in (0,1]$, then we have the strict inequality 
$Q(s_2,a_1)=Q(s_3,a_2)>Q(s_2,a_2)=Q(s_3,a_1)$. For these non-trivial settings of $\gamma$, we can see that the optimal policy for the RL agent is to apply action $a_1$ in state $s_2$ and action $a_2$ in state $s_3$. Lastly, because $s_1$ and $s_4$ are terminal states, the choice of action is irrelevant, as seen in Eq.~\ref{eq:App1}.

The optimal policy is then given by Eq.~\ref{eq:App11}.
\begin{align}
\pi^*(s,a) &= 
\begin{cases} 
  1, &  \text{if $s=2,a_1$ or $s=3,a_2$} \\
  0, &  \text{if $s=2,a_2$ or $s=3,a_1$} \\
  \sfrac{1}{2}, & \text{otherwise}
\end{cases} 
\label{eq:App11}
\end{align}

\section*{Appendix B: Policy Traces and Value}
This section reports the execution traces and corresponding value calculations of a Boolean decision treee with varying $\phi$ for the simple MDP model from Figure \ref{fig:MDP_appendix}.
\begin{center}
\begin{table}
\center
\caption{The set of execution traces for a Boolean decision tree with varying $\phi$, assuming $s_o = 3$. Columns  indicate increasing time, rows indicate settings for $\phi$, and entries indicate $(s_t,R(s_t,a_t),a_t)$.}
\begin{tabular}{lccccc} 
\toprule
$\phi$ &  $t=0$ & $t=1$ & $t=2$ & $t=3$ \\ 
\midrule
$0$    &  $(3,r^+,2)$ & $(2,r^+,2)$ & $(1,r^-,2)$ &  \\ 
$1$    &  $(3,r^+,2)$  & $(2,r^+,2)$ & $(1,r^-,1)$ & \\ 
$2$    &  $(3,r^+,2)$  & $(2,r^+,1)$ & $(3,r^+,2)$ & $(2,r^+,1)$\\ 
$3$    &  $(3,r^+,1)$  & $(4,r^-,2)$ &             &   \\ 
$4$    &  $(3,r^+,1)$  & $(4,r^-,1)$ &             &\\
\bottomrule
\end{tabular}

\label{tab:ValueTraces}
\end{table}
\end{center}

\begin{center}
\begin{table}
\center
\small
\caption{Derived from Table \ref{tab:ValueTraces}, the values $V^{\pi_\phi}$ of Boolean decision tree policies $\pi_{\phi}$ with varying $\phi$ and assuming $s_o = 3$.}
\begin{tabular}{cccccc} 
\toprule
\multirow{2}{*}{$\phi$}  &  \multicolumn{4}{c}{$\gamma^t r^t$} & \multirow{2}{*}{$V^{\pi_{\phi}}(s_3)$} \\ 
\cline{2-5}
 &  $t=0$ & $t=1$ & $t=2$ & $t=3$ &  \\ 
 \midrule
$0$   &  $r^+$ & $r^+\gamma$ & $r^-\gamma^2$ &               & $r^+(1+\gamma) + r^-$ \\ 
$1$   &  $r^+$ & $r^+\gamma$ & $r^-\gamma^2$ &               & $r^+(1+\gamma) + r^-$ \\ 
$2$   &  $r^+$ & $r^+\gamma$ & $r^+\gamma^2$ & $r^+\gamma^3$ & $r^+(1+\gamma+\gamma^2+\gamma^3) $\\ 
$3$   &  $r^+$ & $r^-\gamma$ &              &               & $r^+ + r^-\gamma $ \\
$4$   &  $r^+$ & $r^-\gamma$ &               &               & $r^+ + r^-\gamma $ \\
\bottomrule
\end{tabular}
\label{tab:ValueFunction}
\end{table}
\end{center}

\begin{figure}
  \begin{center}
    \includegraphics[width = 0.3\linewidth]{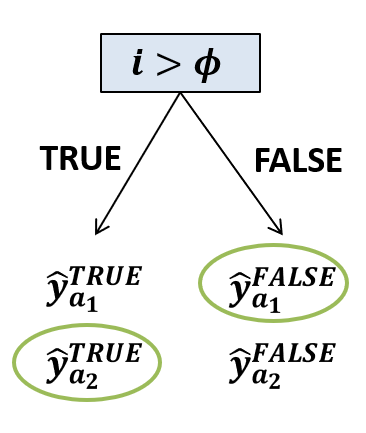}
\end{center}    
    \caption{This figure depicts the tree for our case study.}
    \label{fig:Tree_appendix}
\end{figure}

\section*{Appendix C: Q-learning Leaf Values}
\begin{center}
\begin{table}
\center
\small
\caption{Derived from Table \ref{tab:ValueTraces}, the values $V^{\pi_\phi}$ of Boolean decision tree policies $\pi_{\phi}$ with varying $\phi$ and assuming $s_o = 3$.}
\begin{tabular}{cccc} 
\toprule
Leaf & Q-function & Eq.~& Q-value\\ 
\midrule
$\hat{y}^{FALSE}_{a_2}$ & $Q(s_2,a_2)$ & Eq.~\ref{eq:App4b}&  $r^+ + \gamma r^-$  \\
$\hat{y}^{TRUE}_{a_1}$ & $Q(s_3,a_1)$   & Eq.~\ref{eq:App4c} &  $r^+ + \gamma r^-$\\
$\hat{y}^{FALSE}_{a_1}$ & $Q(s_2,a_1)$  & Eq.~\ref{eq:Rap} &  $ r^+(1+\gamma +\gamma^2 + \gamma^3)$\\
$\hat{y}^{TRUE}_{a_2}$ & $Q(s_3,a_2)$ & Eq.~\ref{eq:Rap}  &  $ r^+(1+\gamma +\gamma^2 + \gamma^3)$ \\
\bottomrule
\end{tabular}
\label{tab:Leaves}
\end{table}
\end{center}

For the decision tree in Fig.~\ref{fig:Tree_appendix}, there are four leaf values: $\hat{y}^{TRUE}_{a_2}$, $\hat{y}^{TRUE}_{a_1}$, $\hat{y}^{FALSE}_{a_2}$, and $\hat{y}^{FALSE}_{a_1}$. Table \ref{tab:Leaves} contains the settings of those parameters. In Table \ref{tab:Leaves}, the first column depicts the leaf parameters; the second column depicts the Q-function state-action pair; the third column contains the equation reference to Appendix A, where the Q-value is calculated; and the fourth column contains the corresponding Q-value. These Q-values assume that the agent begins in a non-terminal state (i.e., $s_2$ or $s_3$) and follows the optimal policy represented by Eq.~\ref{eq:App11}.
% \begin{figure}[h]
%     \includegraphics[width=0.2\textwidth]{images/Tree.png}
%     \caption{}
%     \label{fig:Tree}
% \end{figure}

\newpage 
\section*{Appendix D: Probability of Incorrect Action}
\begin{equation}
\pi_T(s,a) = \mu(s) \hat{y}^{\text{TRUE}}_a + \left(1-\mu(s)\right) \hat{y}^{\text{FALSE}}_a
\label{eq:simpleTree_appendix} 
\end{equation}

\begin{figure}
  \centering
  \includegraphics[width=0.85\linewidth]{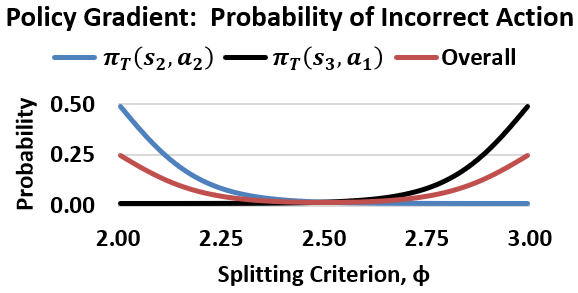}
  \caption{This figure demonstrates the probability of taking a "wrong" action for PG with {\small$\gamma = 0.95$, $a = 10$, $r^+ = 1$, and $r^-=-1$}.}
  \label{fig:WrongAction}
\end{figure}

The output of the differentiable tree is a weighted, nonlinear combination of the leaves (Eq.~\ref{eq:simpleTree_appendix}). Using PG, one samples actions probabilistically from $\pi_T(s,a)$. The probability of applying the ``wrong'' action (i.e., one resulting in a negative reward) is $\pi_T(s_3,a_1)$ in state $s_3$ and $\pi_T(s_2,a_2)$ in state $s_2$. Assuming it equally likely to be in states $s_3$ and $s_2$, the overall probability is $\frac{1}{2}\left(\pi_T(s_2,a_2)+\pi_T(s_3,a_1)\right)$. These probabilities are depicted in Fig.~\ref{fig:WrongAction}, which shows how the optimal setting, $\phi^*$, for $\phi$ should be $\phi^*=2.5$ using PG.

\section*{Appendix E: Architecture Sweeps}
\begin{figure*}
\begin{subfigure}{0.5 \linewidth}
  \centering
  \includegraphics[width=\linewidth]{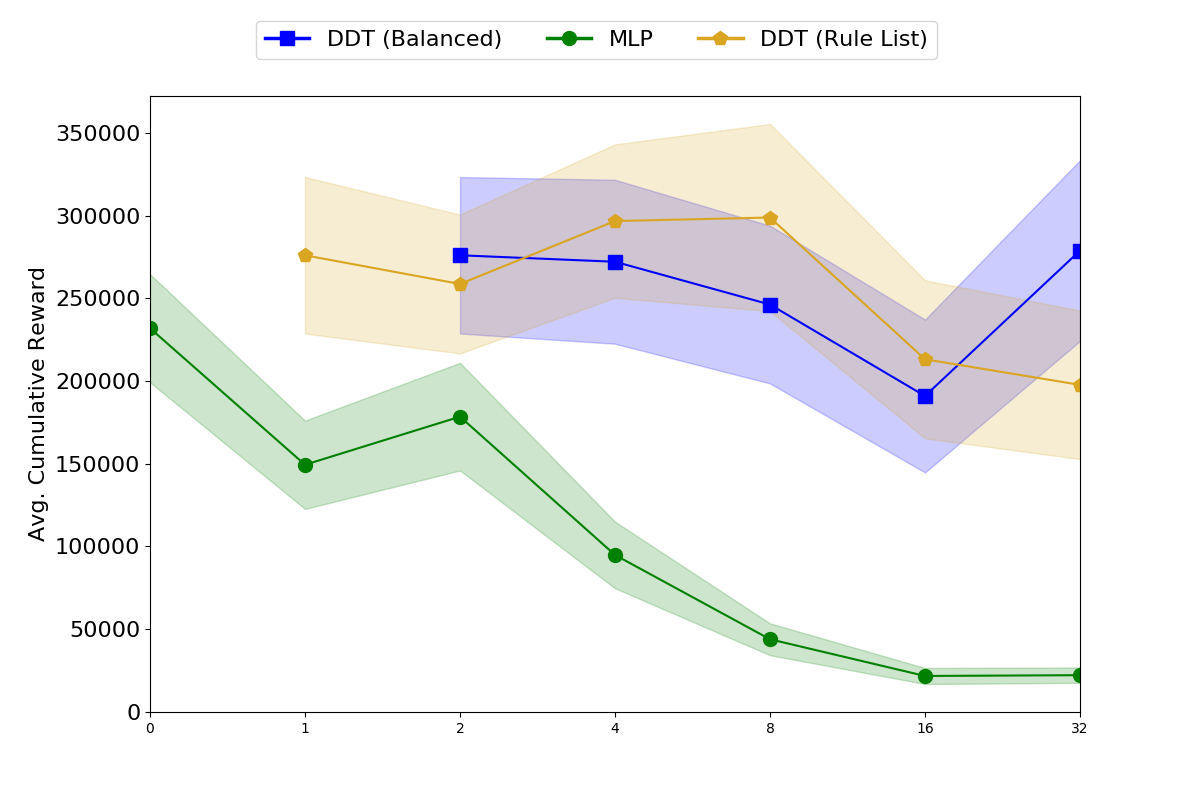}
  \caption{Cart Pole}
  \label{fig:cart-robustness}
\end{subfigure}
\begin{subfigure}{0.5 \linewidth}
  \centering
    \includegraphics[width=\linewidth]{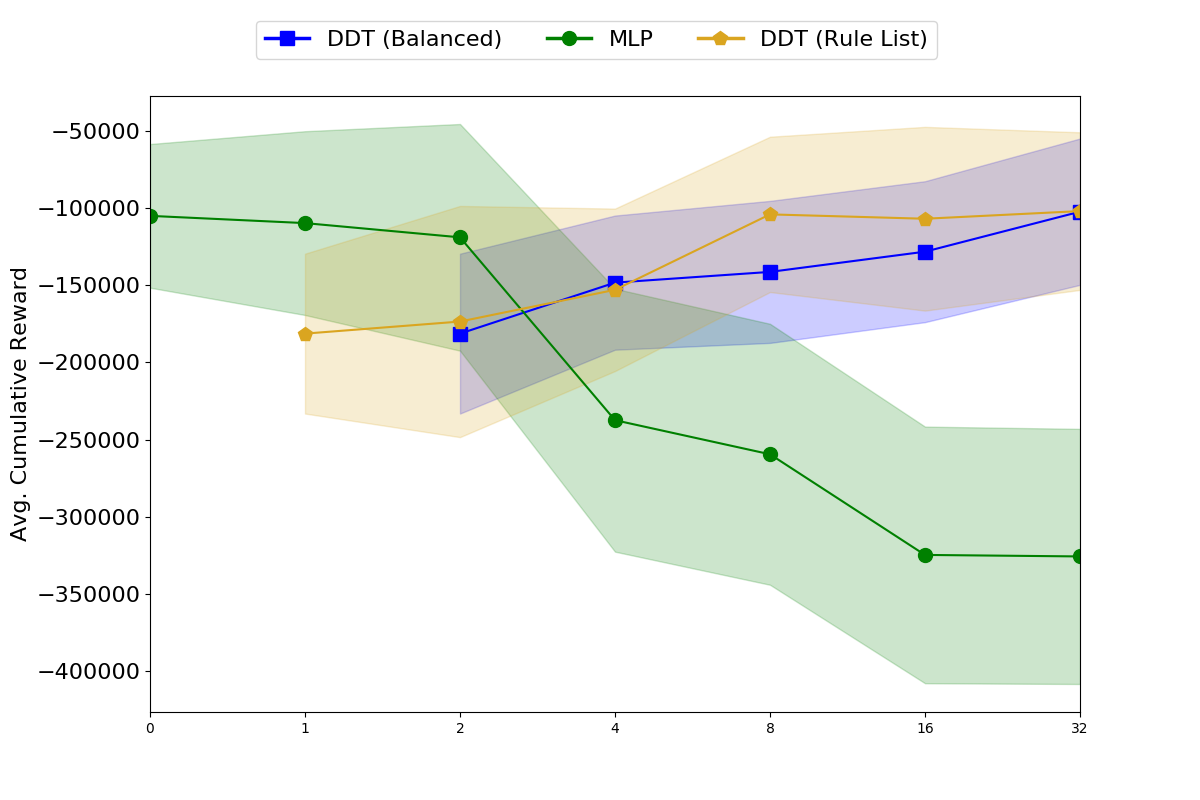}
  \caption{Lunar Lander}
    \label{fig:lunar-robustness}
\end{subfigure}
\caption{Average cumulative reward and standard deviation across architectures of various sizes in the Gym domains. MLP with number of hidden layers, DDT (Rule List) with number of rules, and DDT (Balanced) with number of leaves.}
\label{fig:gym-robustness}
\end{figure*}

\begin{figure*}
\begin{subfigure}{0.5 \linewidth}
  \centering
    \includegraphics[width=\linewidth]{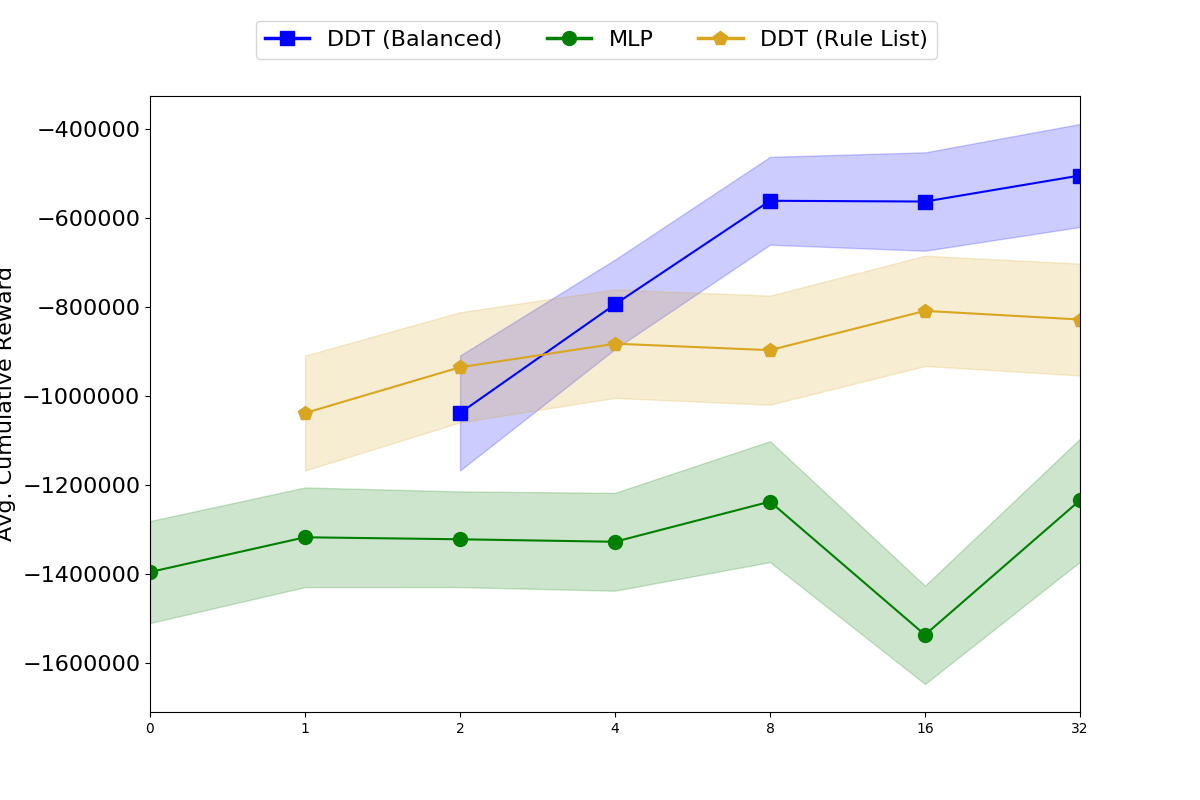}
  \caption{Wildfire Tracking}
    \label{fig:fire-robustness}
\end{subfigure}
\begin{subfigure}{0.5 \linewidth}
  \centering
    \includegraphics[width=\linewidth]{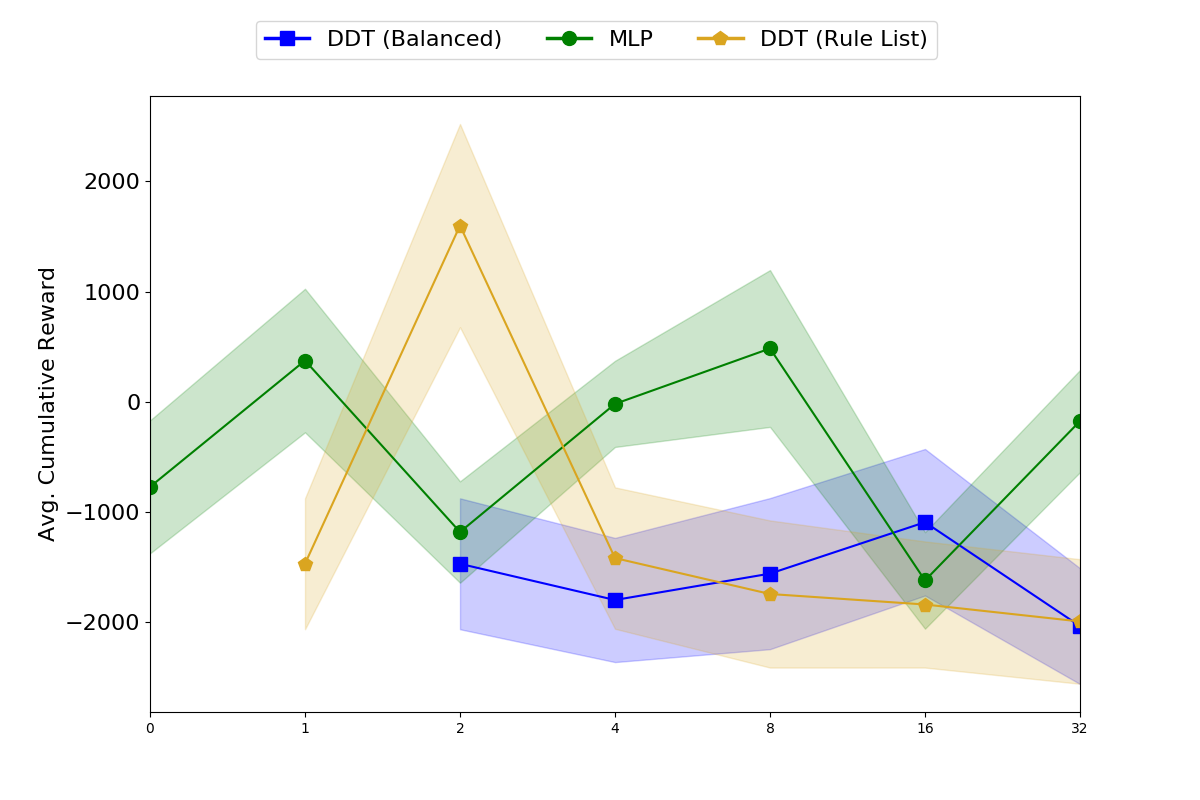}
  \caption{FindAndDefeatZerglings}
    \label{fig:zerg-robustness}
\end{subfigure}
\caption{Average cumulative reward and standard deviation across architectures of various sizes in the wildfire and SC2 domains. MLP with number of hidden layers, DDT (Rule List) with number of rules, and DDT (Balanced) with number of leaves.}
\label{fig:other-robustness}
\end{figure*}

We performed architecture sweeps, as mentioned in the main paper, across all types of models. We found that the MLP requires small models for simple domains, the DDT methods are all relatively unaffected by increased depth, representing a benefit of applying DDTs to various RL tasks. For this result, see Figure \ref{fig:gym-robustness}. As shown in Figure \ref{fig:other-robustness}, in more complex domains, the results are less conclusive and increased depth does not show clear trends for any approach. Nonetheless, we show evidence that DDTs are at least competitive with MLPs for RL tasks of varying complexity, and that they are more robust to hyperparameter tuning with respect to depth and number of layers.

We find that the MLP with no hidden layers performs the best on the two OpenAI Gym domains, cart pole and lunar lander. The best differentiable decision tree architectures for the cart pole domain are those with two leaves and two rules, while the best architectures for lunar lander include 32 leaves and 16 rules.

In the wildfire tracking domain, the 8-layer MLP performed the best of the MLPs, while the 32-leaf differentiable decision tree was the top differentiable decision tree, and the 32-rule differentiable rule list performed the best of the differentiable rule lists.

Finally, the MLP in the FindAndDefeatZerglings domain is an 8-layer MLP, and the differentiable decision tree uses 8 leaves while the differentiable rule list uses 8 rules. 

MLP hidden layer sizes preserve the input data dimension through all hidden layers until finally downsampling to the action space for the final layer. MLP networks all use the ReLU activation after it performed best in a hyperparameter sweep.

\section*{Appendix F: Domain Details}
\subsection{Wildfire Tracking}
\label{subsec:wildfire-results-appendix}

The wildfire tracking domain input space is: \vspace{-8pt}
\begin{itemize}
    \item Fire 1 Distance North (float) \vspace{-8pt}
    \item Fire 1 Distance West (float) \vspace{-8pt}
    \item Closest To Fire 1 (Boolean) \vspace{-8pt}
    \item Fire 2 Distance North (float) \vspace{-8pt}
    \item Fire 2 Distance West (float) \vspace{-8pt}
    \item Closest To Fire 2 (Boolean) \vspace{-8pt}
\end{itemize}
Distance features are floats, representing how far north or west the fire is, relative to the drone. Distances can also be negative, implying that the fire is south or east of the drone.


\subsection{StarCraft II Micro-battle Evaluation}
\label{subsec:sc2-results-appendix}
The FindAndDefeatZerglings manufactured input space is: \vspace{-8pt}
\begin{itemize}
    \item X Distance Away (float) \vspace{-8pt}
    \item Y Distance Away (float) \vspace{-8pt}
    \item Percent Health Remaining (float) \vspace{-8pt}
    \item Percent Weapon Cooldown Remaining (float) \vspace{-8pt}
\end{itemize}
for each agent-controlled unit and 2 allied units, as well as:
\begin{itemize}
    \item X Distance Away (float) \vspace{-8pt}
    \item Y Distance Away (float) \vspace{-8pt}
    \item Percent Health Remaining (float) \vspace{-8pt}
    \item Percent Weapon Cooldown Remaining (float) \vspace{-8pt}
    \item Enemy Unit Type (Boolean) \vspace{-8pt}
\end{itemize}
for the five nearest enemy units. Missing data is filled in with $-1$. The action space for this domain consists of: \vspace{-8pt}
\begin{itemize}
    \item Move North  \vspace{-8pt}
    \item Move East \vspace{-8pt}
    \item Move West \vspace{-8pt}
    \item Move South \vspace{-8pt}
    \item Attack Nearest Enemy \vspace{-8pt}
    \item Attack Second Nearest Enemy \vspace{-8pt}
    \item Attack Third Nearest Enemy \vspace{-8pt}
    \item Attack Second Farthest Enemy \vspace{-8pt}
    \item Attack Farthest Enemy \vspace{-8pt}
    \item Do Nothing \vspace{-8pt}
\end{itemize}

\section*{Interpretable Policies}
\label{sec:policies}
\begin{figure*}
  \centering
  \includegraphics[width=\textwidth]{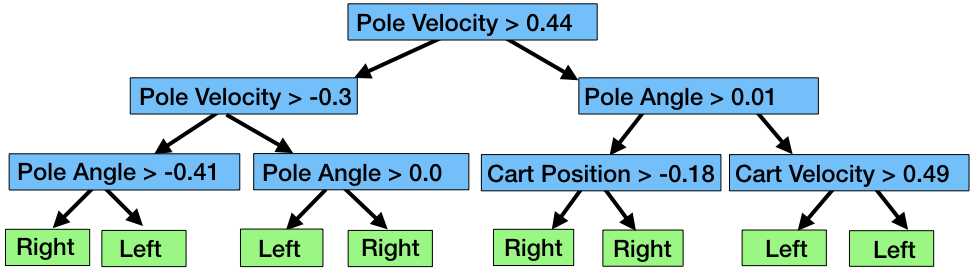}
  \caption{Full interpretable cart pole policy. Two decision nodes are redundant, leading to the same action regardless of how the node is evaluated.}
\label{fig:cart-tree-full}
\end{figure*}

\begin{figure*}
  \centering
  \includegraphics[width=1.1\textwidth,height = 14cm]{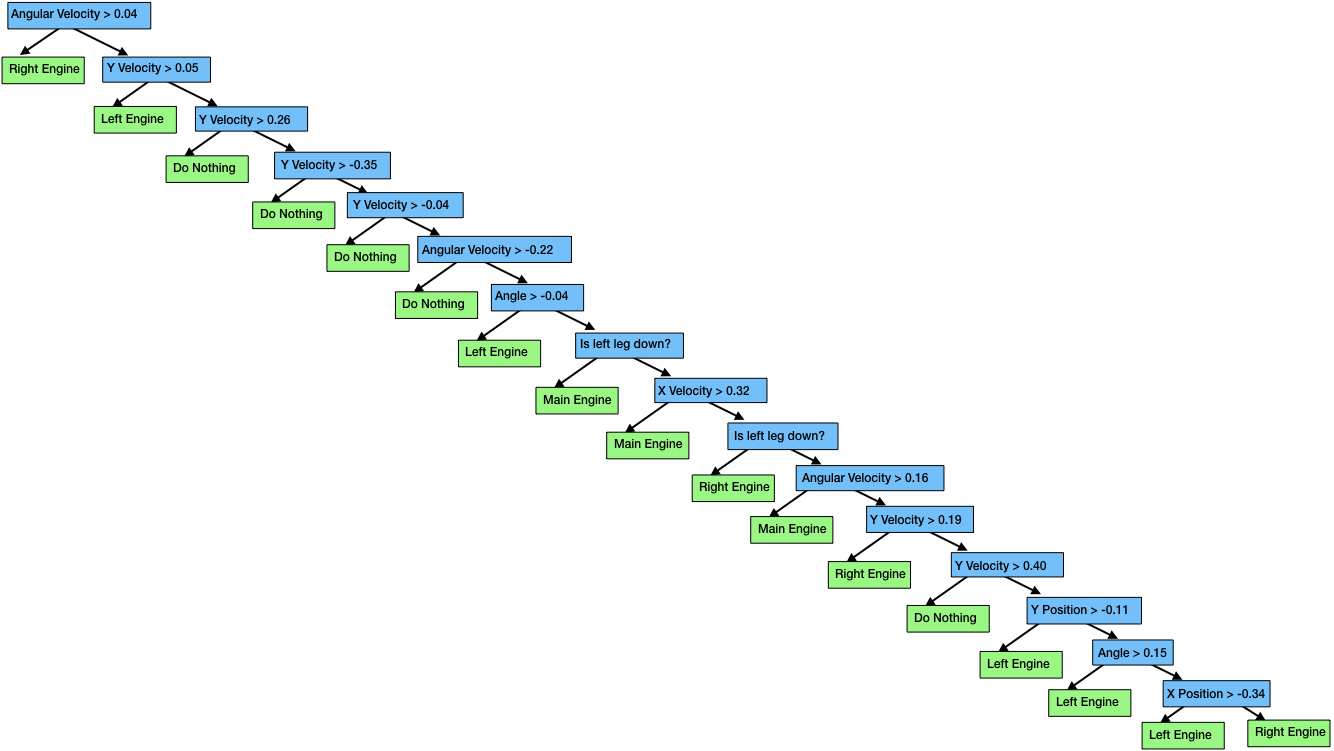}
  \caption{Full interpretable lunar lander rule list policy. Many nodes in the list are not reachable due to previous nodes.}
\label{fig:lunar-tree-full}
\end{figure*}

\begin{figure*}
  \centering
  \includegraphics[width=\textwidth]{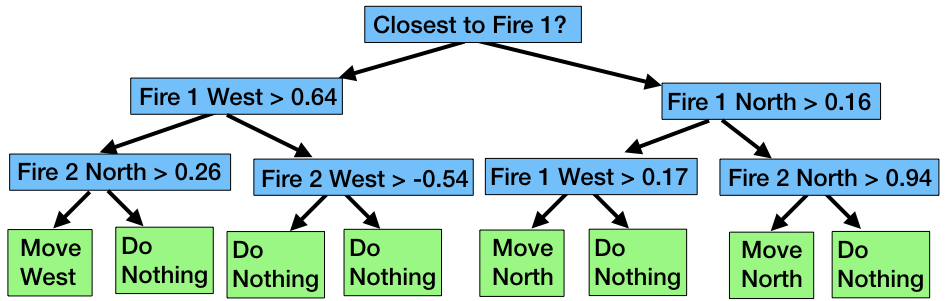}
  \caption{Full interpretable wildfire tracking policy. One node is redundant, leading to the same action regardless of how it is evaluated.}
\label{fig:fire-tree-full}
\end{figure*}

\begin{figure*}
  \centering
  \includegraphics[width=\textwidth]{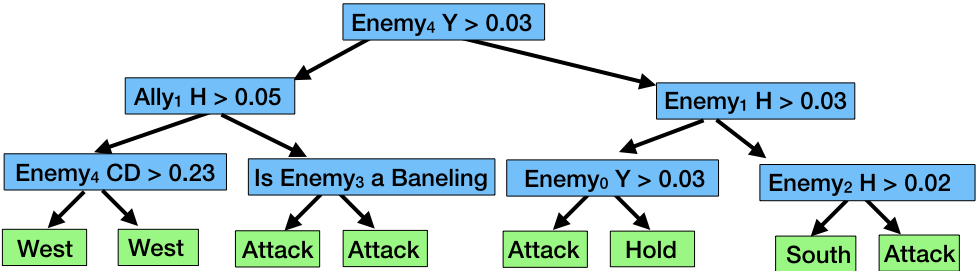}
  \caption{Full interpretable FindAndDefeatZerglings policy. One node is redundant, leading to the same action regardless of how it is evaluated.}
\label{fig:sc2-tree-full}
\end{figure*}

\begin{figure*}
  \centering
  \includegraphics[width=0.75\textwidth]{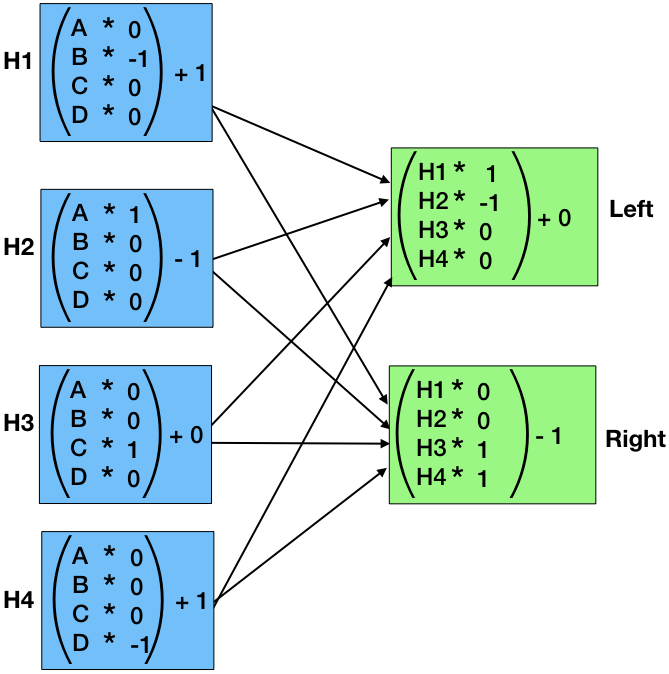}
  \caption{The MLP given to participants for our user study.}
\label{fig:study-mlp-full}
\end{figure*}

\begin{figure*}
  \centering
  \includegraphics[width=0.75\textwidth]{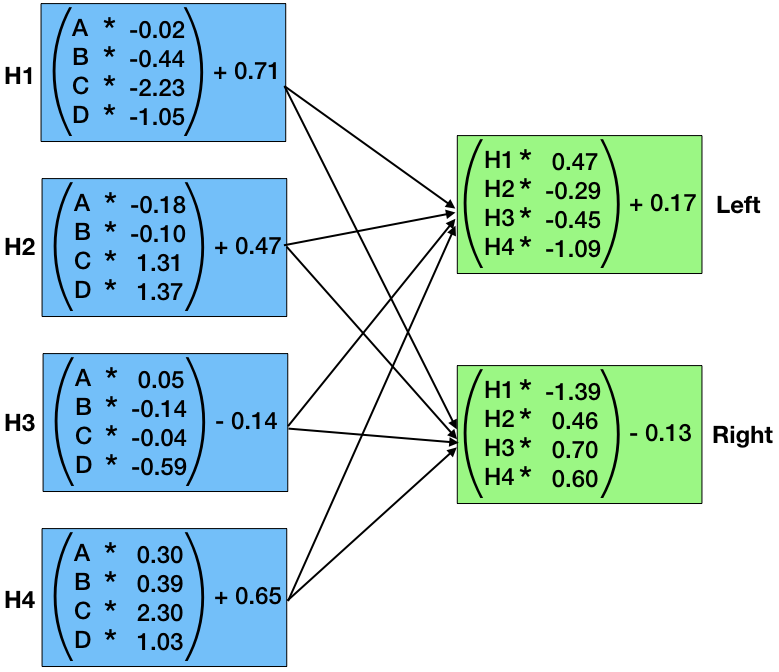}
  \caption{The actual MLP originally intended to go into the user study. Note that it is markedly more complicated than the version given to participants.}
\label{fig:mlp-full}
\end{figure*}

Here we include interpretable policies for each domain, without the pruning that is included in versions in the main body. See Fig.~\ref{fig:cart-tree-full}, \ref{fig:lunar-tree-full}, \ref{fig:fire-tree-full}, and \ref{fig:sc2-tree-full}. Finally, we also include examples of two MLPs represented as decision-making aids. The first is the one-hot MLP that was given to study participants for evaluation of interpretability and efficiency, shown in Figure \ref{fig:study-mlp-full}. The second is the true cart pole MLP, available in Figure \ref{fig:mlp-full}. This decision-making aid turned out to be exceptionally complicated, even with no activation functions and no hidden layer.

\section*{Sample Survey Questions}
\label{sec:questions}
Survey questions included Likert scale questions ranging from 1 (Very Strongly Disagree) to 7 (Very Strongly Agree). For both the MLP and decision trees, some questions included:
\begin{enumerate}
    \item I understand the behavior represented within the model.
    \item The decision-making process does not make sense.
    \item The model's logic is easy to follow
    \item I like the level of readability of this model.
    \item The model is difficult to understand.
\end{enumerate}

\end{document}